\documentclass[10pt,journal]{IEEEtran}
\ifCLASSOPTIONcompsoc
  \usepackage[nocompress]{cite}
\else
  \usepackage{cite}
\fi
\ifCLASSINFOpdf
\else
\fi
\usepackage{epsfig}
\usepackage{graphicx}
\usepackage{caption}
\usepackage{subcaption}
\usepackage{amsmath}
\usepackage{amssymb}
\usepackage{url}
\usepackage{ragged2e}

\usepackage{tikz}
\usetikzlibrary{
    arrows,calc,shapes,positioning,decorations.pathreplacing,
    chains,fit}
\tikzset{
    >=stealth',
}

\graphicspath{{./figures/}{./figures/C-RNNvsSPP-net/SUN_SPP-net/}{./figures/C-RNNvsSPP-net/SUN_C-RNN/}{./figures/C-RNNvsSPP-net/Places_SPP-net/}{./figures/C-RNNvsSPP-net/Places_C-RNN/}}
\DeclareGraphicsExtensions{.pdf,.png,.jpg}

\begin{document}
%
\title{Learning Contextual Dependencies with Convolutional Hierarchical Recurrent\\
	Neural Networks}

\author{Zhen~Zuo,~\IEEEmembership{Student~Member,~IEEE},
        ~Bing~Shuai,~\IEEEmembership{Student~Member,~IEEE}, ~Gang~Wang,~\IEEEmembership{Member,~IEEE},\\
        ~Xiao~Liu,
        ~Xingxing~Wang,
        and~Bing~Wang,~\IEEEmembership{Student~Member,~IEEE}
\IEEEcompsocitemizethanks{\IEEEcompsocthanksitem The authors are with Nanyang Technological University, 50 Nanyang Avenue, Singapore 639798. \protect\\
E-mail: \{zzuo1,bshuai001,wanggang,liux0072,wangxx,wang0775\}@ntu.edu.sg
}}
\IEEEtitleabstractindextext{%
\begin{abstract}
\justifying
Existing deep convolutional neural networks (CNNs) have shown their great success on image classification. CNNs mainly consist of convolutional and pooling layers, both of which are performed on local image areas without considering the dependencies among different image regions. However, such dependencies are very important for generating explicit image representation. In contrast, recurrent neural networks (RNNs) are well known for their ability of encoding contextual information among sequential data, and they only require a limited number of network parameters. General RNNs can hardly be directly applied on non-sequential data. Thus, we proposed the hierarchical RNNs (HRNNs). In HRNNs, each RNN layer focuses on modeling spatial dependencies among image regions from the same scale but different locations. While the cross RNN scale connections target on modeling scale dependencies among regions from the same location but different scales. Specifically, we propose two recurrent neural network models: 1) hierarchical simple recurrent network (HSRN), which is fast and has low computational cost; and 2) hierarchical long-short term memory recurrent network (HLSTM), which performs better than HSRN with the price of more computational cost.  

In this manuscript, we integrate CNNs with HRNNs, and develop end-to-end convolutional hierarchical recurrent neural networks (C-HRNNs). C-HRNNs not only make use of the representation power of CNNs, but also efficiently encodes spatial and scale dependencies among different image regions. On four of the most challenging object/scene image classification benchmarks, our C-HRNNs achieve state-of-the-art results on Places 205, SUN 397, MIT indoor, and competitive results on ILSVRC 2012.
\end{abstract}

}

\maketitle

\IEEEdisplaynontitleabstractindextext

%
\IEEEpeerreviewmaketitle

\vspace{4em}
\IEEEraisesectionheading{\section{Introduction}\label{sec:introduction}}
\begin{figure*}
	\begin{center}
		\includegraphics[width=0.83\linewidth]{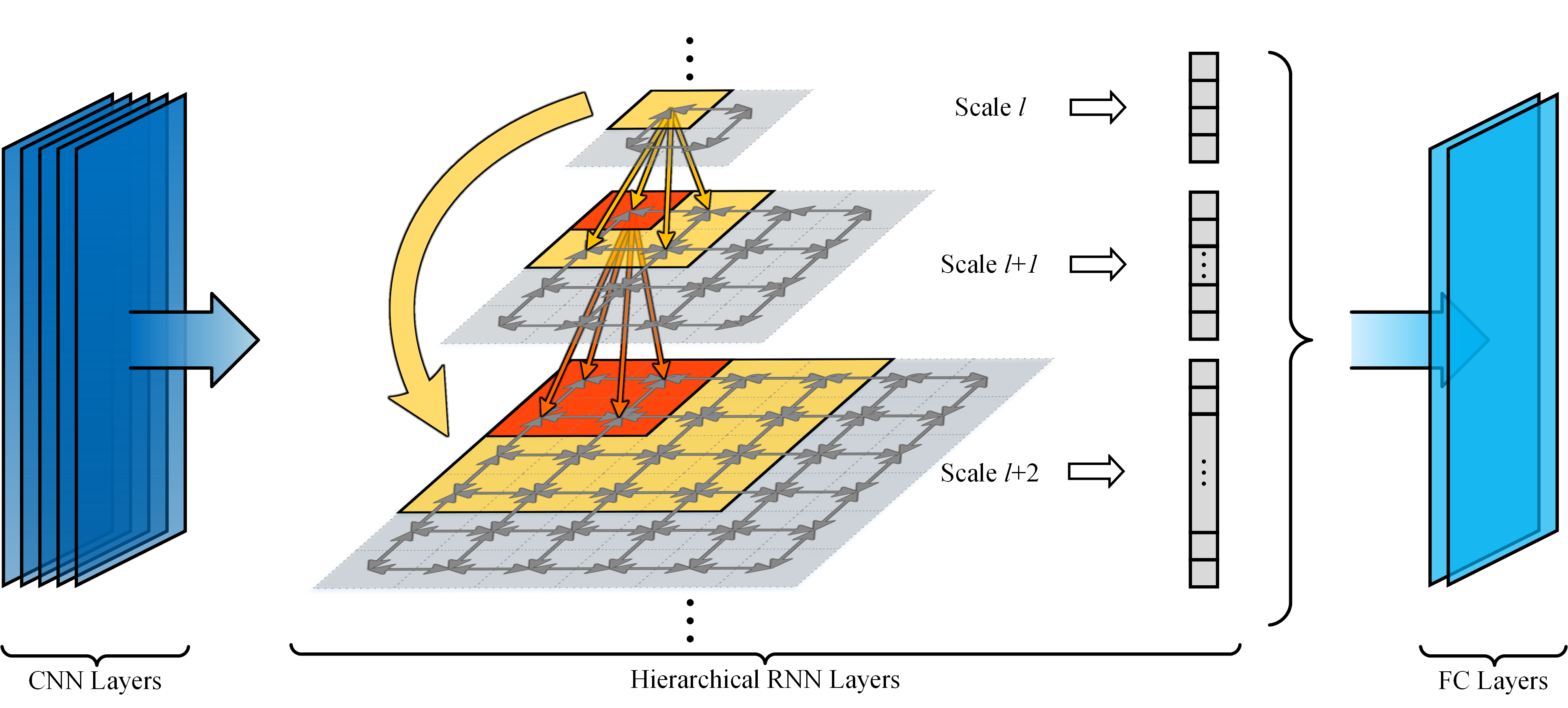}
	\end{center}
	\caption{The overall framework of C-HRNNs. \textbf{CNN layers:} Extract mid-level representations for image regions by processing five convolutional and pooling layers. \textbf{HRNN layers:} \emph{(a)} Pool the output of the fifth CNN layer into multiple scales. \emph{(b)} For each scale, spatial dependencies are captured by direct or indirect connections between each region and its surrounding neighbors. \emph{(c)} For different scales, scale dependencies are encoded by transferring information from the higher level scales to corresponding areas at the lower level scales. Take the $l$-th $(l \in \left[1,\cdots,L\right])$ scale as an example, information of the yellow block will be transferred to corresponding areas (highlighted by yellow) in the $l+1$-th to the $L$-th scales as reference. While the red block in the $l+1$-th scale will be transferred to its influenced areas (highlighted by red) in the $l+2$-th to $L$-th scales. \textbf{FC layers:} Collect different scale HRNN outputs and connect to two fully connected layers. (Best viewed in color)}
	\label{C-HRNN_framework}
\end{figure*}

\IEEEPARstart{O}{ver} the last few years, deep convolutional neural networks \cite{le1990handwritten} have brought a revolution in computer vision society by learning powerful representations based on large-scale datasets. Till now, CNNs have shown their success in but not limited to the following areas: image classification \cite{krizhevsky2012imagenet, zhou2014learning, gong2014multi, chatfield2014return, szegedy2014going, simonyan2014very}, detection \cite{he2014spatial, girshick2014rcnn, sermanet2013overfeat, ouyang2014deepid}, face recognition \cite{taigman2014deepface, sun2014deep}, etc.

The key idea of CNNs is utilizing convolutional and pooling layers to progressively extract more and more abstract patterns. The convolutional layers convolve multiple local filters with input images (or outputs of previous layers), and aim to produce translation invariant local features. Afterwards, pooling layers are applied to summarize the feature responses of the convolutional layers over multiple regions of images, and compress the size of the response maps. Both convolution and pooling are locally performed. For example, the representation of the top left image region will not influence the representation of the bottom right region. However, contextual information is very important for object/scene recognition. For example, in an image with label ``beach", if ``sand" regions are represented with the reference of ``sea" regions, then it is much easier to distinguish them from ``road" or ``desert sand". In CNNs, spatial and scale dependencies among different image regions are not explicitly modeled. 


In this manuscript, we aim to encode contextual dependencies in image representation. To learn the dependencies efficiently and effectively, we propose a new class of hierarchical recurrent neural networks (HRNNs), and utilize the HRNNs to learn such contextual information.

Recurrent neural networks (RNNs) have achieved great success in natural language processing (NLP) \cite{mikolov2012statistical, sutskever2013training, koutnik2014clockwork, graves2014towards, sutskever2009recurrent, boulanger2012modeling}. RNNs \cite{elman1990finding, jaeger2002tutorial} are neural networks developed for modeling dependencies in sequences by using feedback connections among themselves. Thus, they can retain all the processed states in the sequence, and learn patterns from sequential context. Furthermore, because of the reuse of hidden layers, only a limited number of neurons need to be kept in the model. Two most popular RNN models are the simple recurrent neural network (SRN) and long-short term memory recurrent network (LSTM). Based on which, we will introduce hierarchical SRN (HSRN) and hierarchical LSTM (HLSTM). Both HSRN and HLSTM target on modeling the spatial and scale dependencies among different local image regions. However, they also have different characteristics: HSRN is simple and fast, while HLSTM is more complex but it is able to maintain the long-term dependencies among local image regions far away from each other, and lead to better performance than HSRN.

Our proposed hierarchical recurrent neural networks (HRNNs) model two types of contextual dependencies: \emph{spatial dependencies} and \emph{scale dependencies}. 

Firstly, we consider the \emph{spatial dependencies} among image regions from the same scale but at different locations. Since there are no off-the-shelf sequences in images, inspired by the multi-dimensional RNN \cite{graves2009offline}, we generate two dimensional spatial region sequences for images, and represent each region as a function of its neighboring regions. Details will be described in Section \ref{Spatial Dependencies}.

Secondly, we build multiple scale RNNs, and consider \emph{scale dependencies} among image regions from different scales but at the same locations. Information captured from different scales are complementary to each other. Connecting multiple scales can help to learn more robust representation. For example, in an image with label ``car", regions at a lower level scale mostly contain patterns such as ``tire" and ``window", while regions at a higher level scale include global patterns such as ``car". Knowing the existence of ``car" can help the system to increase the representation preference of ``tire" in the corresponding local regions. Details will be described in Section \ref{Scale Dependencies}.

However, HRNN layers are processed based on image regions, while in image classification, no intermediate labels for any of these regions are provided. The only supervision is the image-level labels. To make use of it, fully connected layers are introduced to collect the outputs of HRNN layers, merge them through the global hidden layer, and finally connect to image-level labels with a softmax layer.

Integrating CNNs with our HRNNs, we propose end-to-end networks called convolutional hierarchical recurrent neural networks (C-HRNNs). As shown in Figure \ref{C-HRNN_framework}, C-HRNN not only maintain the discriminative representation power of CNNs, but also efficiently encode the spatial and scale contextual dependencies with HRNNs. Testing on four most challenging large-scale image classification benchmarks, C-HRNNs achieve the state-of-the-arts on Places 205, SUN 397, MIT indoor, and promising results on ILSVRC 2012.


\section{Related Works}
\label{Related Works}
In recent few years, deep neural networks have made great break through in computer vision area. Till now, lots of successful deep neural nets with different structures have been proposed, such as: convolutional neural networks \cite{le1990handwritten, krizhevsky2012imagenet, gong2014multi, chatfield2014return, szegedy2014going, simonyan2014very, he2014spatial, girshick2014rcnn, sermanet2013overfeat, oquab2014learning, donahue2013decaf, taigman2014deepface}, deep belief nets \cite{hinton2006fast,lee2009convolutional, nair20093d}, and auto-encoder \cite{hinton2006reducing, yan2014modeling, zhang2014coarse, bengio2013generalized}, etc. Among all these frameworks, CNNs are the most developed networks for solving image classification problems. The core idea of CNNs is progressively learning more abstract (higher visual level) and more complex patterns: the first few layers focus on learning ``garbor like" low level local features (e.g. edges and lines); based on which, the middle layers target on learning parts of objects (e.g. ``tires" and ``windows" in the images with label ``car"); the higher layers connect to the final image-level labels, and aim to learn representations of the whole image. 

In contrast, RNNs have achieved great success in natural language processing (NLP) \cite{mikolov2012statistical, sutskever2013training, koutnik2014clockwork, graves2014towards, sutskever2009recurrent, boulanger2012modeling, sutskever2011generating, graves2013generating}. Different from the CNNs, which are purely combined with ``feed-forward" network layers, RNNs \cite{elman1990finding, jaeger2002tutorial} are ``feed-back" neural networks designed for modeling contextual dependencies. Because of the connections from the previous states to the current ones, RNNs are networks with ``memory". Through such ``feed-back" connections, RNNs are able to retain information of the past inputs, and it is able to discover correlations among the input data that might be far away from each other in the sequence. 

Although very popular in NLP, RNNs have rarely been applied to computer vision area. In the recent decade, there are mainly five branches of works which involve the recurrent idea.

In the first branch of works, recurrent layers are mainly used as ``tied" layers in the ``feed-forward" networks, which means different layers share the same parameters. Different from our recurrent networks, these ``tied" layers iteratively encode the input data from the same locations with the same network parameters, and these layers focus on reducing the number of parameters, rather than modeling the contextual dependencies among input data from different locations. In \cite{pinheiro2014recurrent}, shared CNNs are applied to learn pixel label consistency among multi-scale image patches. In DrSAE \cite{rolfe2013discriminative}, auto-encoder with rectified linear units are employed to iteratively encode the global digital-number image. In \cite{eigen2013understanding}, the ``tied" CNN (called as recurrent convolutional network) is employed to assess the contributions of the number of layers, response maps, and parameters. Different from these works, the ``recurrent" in our C-HRNNs means learning spatial and scale dependencies among different image regions, and expanding receptive fields of local regions by encoding contextual information.

In the second branch of works, RNNs are used to predict/generate the motion curve of objects/parts in the current/next moment, and applied to visual attention tasks. In \cite{gregor2015draw,mnih2014recurrent}, RNN is used to build a sequential variational auto-encoder to iteratively analyze/generate image parts (at each iteration, RNNs are used to selectively attends to parts of the image while ignoring the others). Differently, we aim to build end-to-end networks for large-scale image classification.

In the third branch of works, RNNs \cite{donahue2014long, ng2015beyond} are used to combine the video information over an ordered sequence of video frames for video recognition and description. Differently, our C-HRNNs model the contextual dependencies within single image rather than the sequential appearance/motion dependencies among consecutive frames.

In the fourth branch of works, RNNs are combined with CNNs for image/video description \cite{chen2014learning, karpathy2014deep, venugopalan2014translating, mao2014deep}. In these works, CNNs are utilized to generate  image/video features, while RNNs are used to connect the image/video feature domain to the text feature domain, and RNNs mainly focus on modeling the text contextual dependencies in the sentences/paragraphs. Different from these works, our C-HRNNs models the contextual information in image appearance domain.

The last branch of works is RNN pyramid \cite{behnke2003hierarchical, visin2015renet}. In these works, multiple layers of local recurrent connectivities are stacked as a pyramid to get different levels of visual abstractions. In contrast, C-HRNNs model the scale dependencies among image regions at the same level of visual abstraction, but different pooling scale. Moreover, C-HRNNs integrate the discriminative power of CNNs and contextual modeling ability of RNNs, and work efficiently and effectively for large-scale image classification.

\begin{figure}
	\centering
	\begin{subfigure}[b]{0.5\textwidth}
		\centering
		\includegraphics[scale=1]{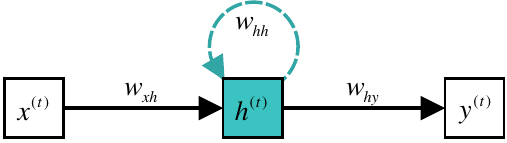}
		\caption{General SRN}
		\label{GeneralSRN}
		\vspace{1em}
	\end{subfigure} 

	~ 
	\centering
	\begin{subfigure}[b]{0.5\textwidth}
		\includegraphics[scale=1]{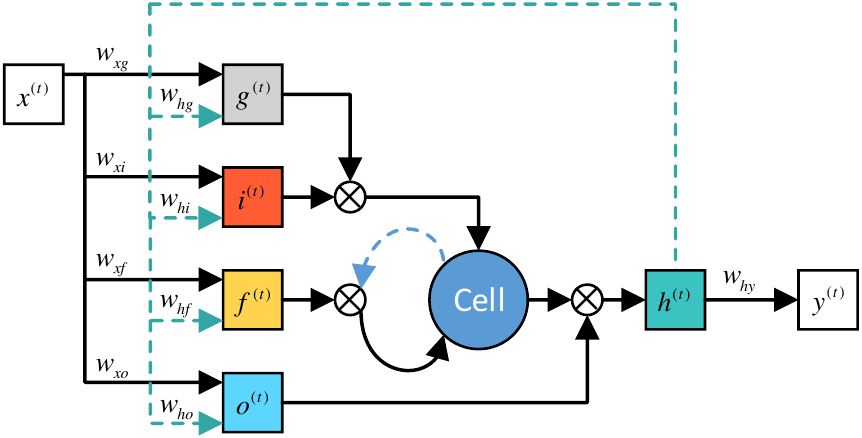}
		\caption{General LSTM}
		\label{GeneralLSTM}
	\end{subfigure}
	\caption{General SRN and LSTM structures, where the solid arrows represent the forward transformations, and the dashed arrows represent the recurrent connections from the previous states to the current ones. (a) SRN structure. In each state $t$ of the sequence, there are two inputs for the hidden layer $h^{(t)}$: the current state input $x^{(t)}$, and the previous state hidden unit $h^{(t-1)}$. The predicted output label $y^{(t)}$ depends on $h^{(t)}$. (b) LSTM structure. Similar to SRN, but the long-term memory can be kept by the introduced intermediate gates (input $i^{(t)}$, forget $f^{(t)}$, output $o^{(t)}$, input modulation $g^{(t)}$ gates), and the memory cell $c^{(t)}$.}\label{fig:animals}
\end{figure}

\section{Convolutional Hierarchical Recurrent Neural Networks}
\label{C-HRNN_algorithm}
As shown in Figure \ref{C-HRNN_framework}, our proposed convolutional hierarchical recurrent neural networks (C-HRNNs) consist of three types of layers: 1) five convolutional (and pooling) layers for extracting middle level image region features; 2) hierarchical recurrent layers for encoding spatial and scale dependencies among different image regions; 3) two fully connected layers for generating global image representation. Finally, an N-way (N indicates the number of categories) softmax loss layer is added on the top for classification. 

\subsection{Convolutional Layers}
\label{Convolutional Neural Network}
As shown in the left part of Figure \ref{C-HRNN_framework}, given input raw pixel images, firstly, five convolutional layers are processed to progressively extract more and more complex and abstract patterns. According to the analysis in \cite{zeiler2013visualizing}, outputs of the fifth convolutional layer are able to capture patterns representing parts and objects. Furthermore, size of the fifth layer response maps is orders of magnitudes smaller than size of the original raw pixel images. Thus, based on such CNN features, our proposed HRNNs can model the contextual dependencies among middle-level regions with semantic meanings, and HRNNs can be processed very efficiently. Furthermore, with back propagation, RNNs can help the CNNs to increase the quality of middle-level and low-level features.

Note that our HRNNs can be easily constructed based on any network other than CNN (e.g. deep restricted Boltzmann machine \cite{sutskever2009recurrent}, auto-encoder \cite{rolfe2013discriminative}), hand crafted features (e.g. SIFT \cite{lowe2004distinctive}, HOG \cite{dalal2005histograms}), or even from scratch. In this work, we choose CNNs because of their excellent performance on representing mid-level patterns, which is the guarantee of good performance of the following HRNNs.

\subsection{Review of General RNNs}
\label{General RNNs}
RNNs \cite{elman1990finding, jaeger2002tutorial} are originally developed for modeling dependencies in time sequential data. In RNNs, two of the most typical models are the simple recurrent neural network (SRN), and the long-short term memory recurrent neural network (LSTM). In the following two subsections, SRN and LSTM will be introduced to represent each state $t$ of a given sequence of length $T$. $x^{(t)}$, $h^{(t)}$, and $y^{(t)}$ are the input, hidden and output representations of the $t$-th state respectively. 

\subsubsection{Simple Recurrent Neural Nets}
As shown in Figure \ref{GeneralSRN}, the $t$-th state in SRN can be represented as:
\begin{align}
	{h^{(t)}} & = \psi_h\left( W_{hh}h^{{({t-1})}}+W_{xh}x^{(t)}+b_h\right) \label{RNN_basic_form}\\
	{y^{(t)}} & = \psi_y\left( W_{hy}h^{(t)}+b_y\right) \label{RNN_basic_form_output}
\end{align}
where $W_{xh}$, $W_{hh}$ and $W_{hy}$ are the shared transformation matrices from input to hidden states, previous hidden to current hidden states, and hidden to output states. $b_h$ and $b_y$ are bias terms, $\psi_h$ and $\psi_y$ are non-linear activation functions. Since the expression of each state is based on hidden representation of the previous states, SRN can keep ``memory" of the whole sequence, and learn patterns based on such sequential context.

Although simple and effective, SRN has the unpleasant ``short term memory" problem \cite{hochreiter1997long}: during the back-propagation procedure in SRN, the gradients will be multiplied $T$ times by the $W_{hh}$. Consequently, when $T$ is relatively large, there will be gradient vanishing/exploding problems.

\subsubsection{Long-Short Term Memory Recurrent Neural Nets}
To overcome the above ``short term memory" issue, LSTM \cite{hochreiter1997long} introduce the ``memory block" (combined with multiplication gates and memory cell) to keep long term flow of sequential information. As shown in Figure \ref{GeneralLSTM}, the $t$-th state in LSTM can be represented as:
\begin{align}
	{i^{(t)}} & = \sigma\left(W_{hi}h^{{({t-1})}} + W_{xi}x^{(t)} + b_i\right)\\
	{f^{(t)}} & = \sigma\left(W_{hf}h^{{({t-1})}} + W_{xf}x^{(t)} + b_f\right)\\
	{o^{(t)}} & = \sigma\left(W_{ho}h^{{({t-1})}} + W_{xo}x^{(t)} + b_o\right)\\
	{g^{(t)}} & = \phi\left(W_{hg}h^{{({t-1})}} + W_{xg}x^{(t)} + b_g\right)\\
    {c^{(t)}} & = {f^{(t)}}\odot {c^{(t-1)}} + {i^{(t)}}\odot {g^{(t)}}
	\label{LSTM_basic_form_c}\\
	{h^{(t)}} & = {o^{(t)}}\odot \phi\left({c^{(t)}}\right) \\
	{y^{(t)}} & = \psi_y\left( W_{hy}h^{(t)}+b_y\right)
	\label{LSTM_basic_form_output}
\end{align}
in addition to the hidden state $h^{(t)}$, LSTM introduced a memory cell $c^{(t)}$, and four multiplication gates: $i^{(t)}$, $f^{(t)}$, $o^{(t)}$, and $g^{(t)}$, which are the input, forget, output, and input modulation gate respectively. $\sigma$ is a logistic sigmoid function (thus, $i^{(t)}$, $f^{(t)}$, $o^{(t)}$ range from $[0,1]$). $\phi$ is the hyperbolic tangent nonlinearity, and $\odot$ represents element-wise multiplication. Specifically, the self recurrent memory cell $c^{(t)}$ keeps the long-term memory. The input gate $i^{(t)}$ controls the flow of incoming signal to alter the state of $c^{(t)}$. The forget gate $f^{(t)}$ helps the $c^{(t)}$ to selectively maintain and forget the previous state status $c^{(t-1)}$. While the output gate $o^{(t)}$ controls the amount of memory that transmits to $h^{(t)}$. 

The ``memory block" structure enables LSTM to selectively forget its previous memory states, and learn long-term dynamics which general SRN can hardly handle. However, LSTM has more intermediate neurons than SRN, thus LSTM consumes much more computational resources.

In this manuscript, rather than modeling contextual correlations among different states in time sequences, we modify RNNs to model contextual dependencies among image region ``2D sequences". Details of our proposed networks will be introduced in the following section.

\subsection{Hierarchical Recurrent Layers}
\label{Recurrent Neural Network}
In CNNs, convolution and pooling are locally performed on image regions. While the spatial dependencies among different regions from the same scale are ignored, let alone the scale dependencies among image regions from different scales. On the other hand, general RNNs (Section \ref{General RNNs}) are designed for modeling dependencies in sequences, however, they cannot be directly applied on images. Thus, as shown in the middle part of Figure \ref{C-HRNN_framework}, we propose hierarchical recurrent layers to model spatial and scale contextual dependencies.

\subsubsection{Modeling Spatial Contextual Dependencies}
\begin{figure*}
	\begin{center}
		\includegraphics[width=0.95\linewidth]{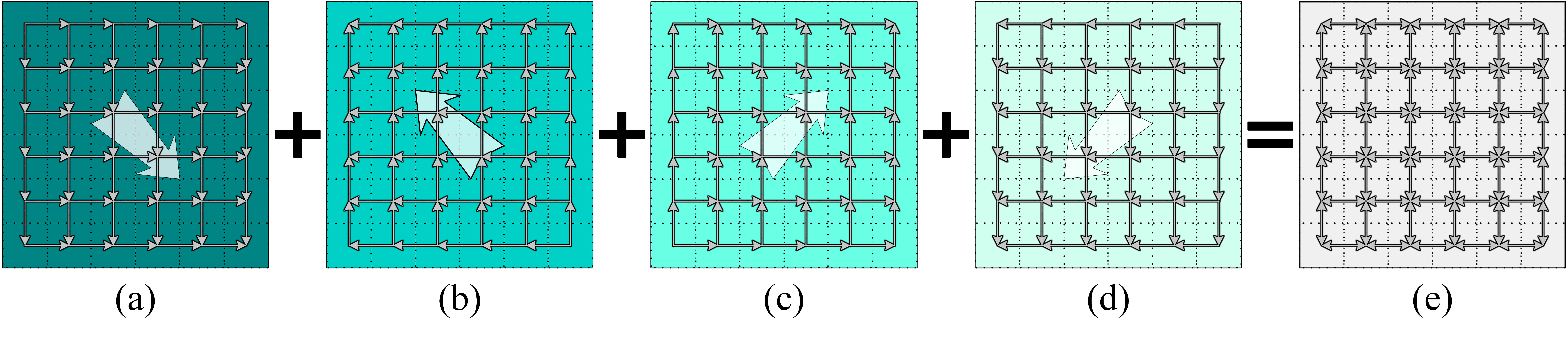}
	\end{center}
	\caption{Overview of a single scale HRNN layer ($6\times{}6$ image regions, one dashed box corresponds to one image region). \emph{(a)}: Transmit information from left and top spatial neighbors for each region, and process from top-left corner to bottom-right corner in an acyclic way. \emph{(b-d)}: Similar to \emph{(a)}, except the processing directions are bottom-right to top-left, bottom-left to top-right, and top-right to bottom-left respectively.  \emph{(e)}: Combine \emph{(a-d)}, then each image region has direct or indirect contextual references from all the other regions. (Best viewed in color)}
	\label{spatialRNN}
\end{figure*}

\label{Spatial Dependencies}
Spatial context is an important clue for recognizing images. For example, in an image with label ``computer room", knowing the existence of ``computer" can help the system to increase the preference of representing ``desk" in the surrounding image regions. In this subsection, we will introduce the spatial RNNs to model spatial contextual dependencies within single scale image feature maps.

There are no existing sequences in images, hence we need to generate region sequences in image domain. Take Alex-net \cite{krizhevsky2012imagenet} as an example, as described in Section \ref{Convolutional Neural Network}, we utilize the fifth layer CNN feature maps ($256\times{}6\times{}6$, corresponding to number of channels $\times$ height $\times$ width) as the input of the recurrent layers. It can be considered as a $6\times{}6$ 2D data array, each element in the array is represented as a 256 dimensional vector. Then how to convert such an 2D array into sequences? The most straight-forward way is to scan in a row by row or column by column manner. However, images are 2-dimensional data. For each element, contextual information from all the directions should be taken into consideration. Thus, inspired by \cite{graves2009offline}, we generate ``2D sequences" for images, and each element simultaneously receives spatial contextual references from its 2D neighborhood elements. 

As shown in \emph{(e)} of Figure \ref{spatialRNN}, spatial contextual information comes from all directions (left, right, top, bottom). If we directly connect all the surrounding elements to the target, each node would simultaneously be the ``previous" and ``next" element of its neighbors. Then the connections would form a cyclic graph. The resulting network is difficult to be optimized. Thus, four directional ``2D sequences" are generated for each scale: top-left to bottom-right, bottom-right to top-left, bottom-left to top-right, and top-right to bottom-left. Each of them focuses on transferring information from an independent direction through an acyclic path. Take the top-left to bottom-right sequence (as shown in \emph{(a)} of Figure \ref{spatialRNN}) as an example, each element receives references from its nearest neighbor elements in the previous row and the previous column. All the elements will be visited once, and each element can be unrolled into a function of all the previously visited elements. Similarly, contextual information from the other three directions can be encoded by ``2D sequences" as shown in \emph{(b-d)} of Figure \ref{spatialRNN}. 

For each of the four directional ``2D sequences", the transformation matrices are shared through the whole sequence. To model the spatial correlations among different image regions, general SRN and LSTM (Section \ref{General RNNs}) are modified to model our spatial sequences, and they are called spatial SRN and spatial LSTM in the rest of this section.

\vspace{1ex}
\noindent \textbf{Spatial SRN} Firstly, the spatial SRN is introduced. The hidden representation of each image region in the ``2D sequences" is:
\begin{align}
\nonumber
	h^{(r, c)}_\searrow & = \psi_{h}^{}\Big(W_{hh\searrow}^{r} h^{({r-1, c})}_\searrow+W_{hh\searrow}^{c} h^{({r, c-1})}_\searrow\\
	& ~~~~~+W_{x{h\searrow}}^{}x^{(r,c)}+b_{h\searrow}^{}\Big)
	\label{RNNquad_form_se}\\
\nonumber
	h^{(r, c)}_\nwarrow & = \psi_{h}^{}\Big(W_{hh\nwarrow}^{r} h^{({r+1, c})}_\nwarrow+W_{hh\nwarrow}^{c} h^{({r, c+1})}_\nwarrow\\
	& ~~~~~+W_{x{h\nwarrow}}^{}x^{(r,c)}+b_{h\nwarrow}^{}\Big)  \label{RNNquad_form_nw}\\
\nonumber
	h^{(r, c)}_\nearrow & = \psi_{h}^{}\Big(W_{hh\nearrow}^{r} h^{({r+1, c})}_\nearrow+W_{hh\nearrow}^{c} h^{({r, c-1})}_\nearrow\\
	& ~~~~~+W_{x{h\nearrow}}^{}x^{(r,c)}+b_{h\nearrow}^{}\Big) 
	\label{RNNquad_form_ne}\\
\nonumber
	h^{(r, c)}_\swarrow & = \psi_{h}^{}\Big(W_{hh\swarrow}^{r} h^{({r-1, c})}_\swarrow+W_{hh\swarrow}^{c} h^{({r, c+1})}_\swarrow\\
	& ~~~~~+W_{x{h\swarrow}}^{}x^{(r,c)}+b_{h\swarrow}^{}\Big)
	\label{RNNquad_form_sw}\\	
	{h^{(r, c)}} & = {h^{(r, c)}_\searrow} + {h^{(r, c)}_\nwarrow} + {h^{(r, c)}_\nearrow} + {h^{(r, c)}_\swarrow}  \label{RNNquad_form}
\end{align}
where $(r,c)$ is the position of the element. $x^{(r,c)}$ is the input, which is an image region represented by a 256-dimensional fifth CNN layer feature vector, ${h^{(r,c)}_\searrow}$, ${h^{(r,c)}_\nwarrow}$, ${h^{(r,c)}_\nearrow}$, and ${h^{(r,c)}_\swarrow}$ denote the hidden representations of $x^{(r,c)}$ in the four ``2D sequences" respectively (corresponding to top-left to bottom-right, bottom-right to top-left, bottom-left to top-right, and top-right to bottom-left directions).  $h^{(r,c)}$ is the combination of the four directional hidden representations, which is the output of spatial SRN. For each direction, $W_{hh}^{r}$ and $W_{hh}^{c}$ are row based and column based hidden to hidden states transformation matrices. $W_{xh}$ is the input to hidden states transformation matrix, $b_h$ is the bias term, and $\psi_h$ is a non-linear activation function (ReLU is used here).

\vspace{1ex}
\noindent \textbf{Spatial LSTM} Similar to the Equation \ref{RNNquad_form} in spatial SRN, each hidden unit $h^{(r, c)}$ of spatial LSTM is also a combination of four directional hidden representations. To make the expression concise, only functions corresponding to the $\searrow$ direction will be expanded here (corresponding to Equation \ref{RNNquad_form_se}):
\begin{align}
\nonumber
	{i^{(r, c)}_\searrow} & = \sigma\Big(W_{hi\searrow}^{r}h^{{({r-1, c})}}_\searrow + W_{hi\searrow}^{c}h^{{({r, c-1})}}_\searrow \\
	& ~~~~~+ W_{xi\searrow}x^{(r,c)} + b_{i\searrow}\Big)
	\label{LSTMquad_i}\\
\nonumber
	{f^{(r, c)}_\searrow} & = \sigma\Big(W_{hf\searrow}^{r}h^{{({r-1, c})}}_\searrow +
	W_{hf\searrow}^{c}h^{{({r, c-1})}}_\searrow \\
	& ~~~~~+ W_{xf\searrow}x^{(r,c)} + b_{f\searrow}\Big)
	\label{LSTMquad_f}\\
\nonumber
	{o^{(r, c)}_\searrow} & = \sigma\Big(W_{ho\searrow}^{r}h^{{({r-1, c})}}_\searrow +
	W_{ho\searrow}^{c}h^{{({r, c-1})}}_\searrow \\
	& ~~~~~+ W_{xo\searrow}x^{(r,c)} + b_{o\searrow}\Big)
	\label{LSTMquad_o}\\
\nonumber
	{g^{(r, c)}_\searrow} & = \phi\Big(W_{hg\searrow}^{r}h^{{({r-1, c})}}_\searrow +
	W_{hg\searrow}^{c}h^{{({r, c-1})}}_\searrow \\
	& ~~~~~+ W_{xg\searrow}x^{(r,c)} + b_{g\searrow}\Big)
	\label{LSTMquad_g}\\		
    {c^{(r, c)}_\searrow} & = {f^{(r, c)}_\searrow} \odot \Big({c^{(r-1, c)}_\searrow} + 
    {c^{(r, c-1)}_\searrow}\Big) + {i^{(r, c)}_\searrow}\odot {g^{(r, c)}_\searrow}
    \label{LSTMquad_c}\\
	{h^{(r, c)}_\searrow} & = {o^{(r, c)}_\searrow} \odot \phi\left({c^{(r, c)}_\searrow}\right)
	\label{LSTMquad_h}
\end{align}
where $(r,c)$ is the current state position, $x^{(r,c)}$ represents the current input data. $i^{(r,c)}_\searrow$, $f^{(r,c)}_\searrow$, $o^{(r,c)}_\searrow$, $g^{(r,c)}_\searrow$, correspond to the input, forget, output, and input modulation gates. $c^{(r,c)}_\searrow$ denotes the memory cell unit, and finally ${h^{(r,c)}_\searrow}$ is the hidden representations of $x^{(r,c)}$ in top-left to bottom-right direction. Similarly, the other three directions can be achieved.

For each gate function, $W_{hp}^{r}$, $W_{hp}^{c}$, $W_{xp}^{c}$ and $b_p$ ($p\in{\{i, f, o, g\}}$) are hidden-gate (row), hidden-gate (column), input-gate transformation matrices and bias terms respectively. $\sigma$ and $\phi$ are non-linear activation functions, in this manuscript, sigmoid is used as $\sigma$, and tangent is assigned as  $\phi$.

\subsubsection{Modeling Scale Contextual Dependencies}
\label{Scale Dependencies}
Besides spatial contextual dependencies, there also exist scale contextual dependencies among image regions from the same locations but at different scales, which is another important clue for image recognition. For example, again in an image with label ``computer room", knowing the global pattern ``computer room" can help the system to increase the preference of representing patterns correspond to ``computer" and ``desk" in local level scales. In this subsection, we will focus on modeling scale dependencies.

The final goal of image classification is to achieve good image-level representation, which is based on well-performed local image region representations. When describing a local image region, the traditional way is to only encode its own information. In contrast, if information from higher level scale regions is given, then the global information would be encoded in local features, and lead to better local descriptions. Thus, we build connections across regions from different scales.

For each element at each scale, its receptive field covers a number of elements at the lower level scales. More intuitively, as shown in the middle part of Figure \ref{C-HRNN_framework}, areas highlighted with yellow at the scale $l+1$ and $l+2$ are covered by the receptive field of the yellow element at the scale $l$. Thus, global information from the higher level scale $l$ would be transferred to the corresponding areas at the lower level scales $l+1$ and $l+2$. Thus, for element at position $(r_l, c_l)$  on scale $l$, the scale dependencies from higher level scales can be encoded as:
\begin{equation}
s_{l}^{(r_l,c_l)} = \sum_{j=1}^{l-1}~W_{jl}h_j^{(r_j,c_j)}
\label{RNN_scale}
\end{equation}
where $l \in \left[2, \cdots, L\right]$, and $L$ is the number of scales. $(r_j,c_j)$ is the position at the higher level scale $j$. $h_j^{(r_j,c_j)}$ is scale contextual element (already combined the four directional spatial dependencies, refer to Equation \ref{RNNquad_form}) from the higher level scale. $W_{jl}$ is the scale $j$ to scale $l$ transformation matrix. 

\subsubsection{HRNNs with Spatial \& Scale Dependencies}
By inserting Equation \ref{RNN_scale} into Equation \ref{RNNquad_form_se}-\ref{RNNquad_form_sw} (spatial SRN) or Equation \ref{LSTMquad_i}-\ref{LSTMquad_h} (spatial LSTM), scale and spatial dependencies can be modeled together in our hierarchical RNNs. 

\vspace{1ex}
\noindent \textbf{HSRN} Take the top-left to right-bottom directional HSRN as an example (refer to Equation \ref{RNNquad_form_se}), the hidden representation of each element is:
\begin{align}
\nonumber
\tilde{h}^{(r, c)}_\searrow & = \psi_{h}^{}\Big(W_{hh\searrow}^{r} \tilde{h}^{({r-1, c})}_\searrow+W_{hh\searrow}^{c} \tilde{h}^{({r, c-1})}_\searrow\\
& ~~~~~+s^{(r,c)}+W_{x{h\searrow}}^{}x^{(r,c)}+b_{h\searrow}^{}\Big) \label{HRNNquad_form_se}\\
\nonumber
{s^{(r,c)}} & = \sum_{j=1}^{l-1}~W_{jl}\tilde{h}_j^{(r_j,c_j)}
\end{align}

\vspace{1ex}
\noindent \textbf{HLSTM} Similarly, for the HLSTM model (refer to Equation \ref{LSTMquad_i}-\ref{LSTMquad_g}), the hidden representation of each gate functions is:
\begin{align}
\nonumber
	{\tilde{p}^{(r, c)}_\searrow} &= \sigma\Big(W_{hp\searrow}^{r}\tilde{h}^{{({r-1, c})}}_\searrow + W_{hp\searrow}^{c}\tilde{h}^{{({r, c-1})}}_\searrow \\
	& ~~~~~+s^{(r,c)}+W_{xp\searrow}x^{(r,c)} + b_{i\searrow}\Big)
	\label{HLSTMquad_form_se}\\
\nonumber
{s^{(r,c)}} & = \sum_{j=1}^{l-1}~W_{jl}\tilde{h}_j^{(r_j,c_j)}	\\	
\nonumber	
\tilde{p} &\in{\{\tilde{i}, \tilde{f}, \tilde{o}, \tilde{g}\}}
\end{align}
in which, $\sigma$ is sigmoid function when $\tilde{p} \in{\{\tilde{i}, \tilde{f}, \tilde{o}\}}$, and $\sigma$ is tangent when $\tilde{p} = \tilde{g}$.

For both Equation \ref{HRNNquad_form_se} and \ref{HLSTMquad_form_se}, the scale index $l$ of each variable is removed for the convenience of expression. Similarly, expressions of the other three directions can be obtained. Afterwards, refer to Equation \ref{RNNquad_form}, by combining the revised four directional hidden representations, the complete HRNNs (for both HSRN and HLSTM) hidden element expression is:
\begin{equation}
{\tilde{h}^{(r, c)}}  = {\tilde{h}^{(r, c)}_\searrow} + {\tilde{h}^{(r, c)}_\nwarrow} + {\tilde{h}^{(r, c)}_\nearrow} + {\tilde{h}^{(r, c)}_\swarrow} 
\label{HRNNquad_form}
\end{equation}

According to the RNNs optimization notes in \cite{mikolov2012statistical}, RNNs can be simply and effectively optimized by back propagation through time (BPTT). In BPTT, the recurrent nets would be unfolded into feed-forward deep networks, then normal back-propagation can be applied. Utilizing the ``weight sharing" setting in Caffe \cite{jia2014caffe}, BPTT can be performed with shared RNN weights. 

\subsection{Fully Connected Layers}
Different from applications like image labeling, where the label $y^{(r,c)}$ of each pixel or patch level image region $x^{(r,c)}$ is given, in image classification, there is no intermediate labels except the overall image-level label. Thus, Equation \ref{RNN_basic_form_output} (corresponds to HSRN) or Equation \ref{LSTM_basic_form_output} (corresponds to HLSTM) cannot be directly applied. To connect the hierarchical recurrent layers (Equation \ref{HRNNquad_form}) to the image labels, fully connected layers are introduced to merge the information learned by different scales of HRNNs:
\begin{align}
	\label{RNNbi_form_output}
	g & =  \psi_g(W_{hg}H+b_g) \\
	y & =  \varphi_o(W_{go}g+b_o) \\
\nonumber
	{\rm{where}}\hspace{3pt} H & = [\hspace{2pt} \tilde{h}_1^T, \cdots, \tilde{h}_l^T, \cdots, \tilde{h}_L^T \hspace{2pt}]^T\\
\nonumber
	{\rm{and}}\hspace{11pt}  \tilde{h}_l &= [(\tilde{h}_l^{(1,1)})^T, \cdots, (\tilde{h}_l^{(r_l,c_l)})^T, \cdots, (\tilde{h}_l^{(R_l,C_l)})^T]^T
\end{align}
where $W_{hg}$ is the fully connected transformation matrix to transform the HRNNs output $H$ to the global hidden layer $g$. $H$ is the concatenation of HRNN layer outputs $\tilde{h}_l$ ($l\in \left[1, \cdots, L\right]$) at different scales. For each scale, $\tilde{h}_l$ is the concatenation of all its hidden element expressions $\tilde{h}_l^{(r_l,c_l)}$ ($r_l\in \left[1, \cdots, R_l\right]$, $c_l\in \left[1, \cdots, C_l\right]$, $R_l$ and $C_l$ are the number of rows and columns at the scale $l$ respectively). $W_{go}$ is learned to connect $g$ with the class label $y$, $b_g$ and $b_o$ are the bias terms. $\psi_g$ is a non-linear activation function (ReLU is used in this manuscript), and $\varphi_o$ is the softmax function for classification.


\section{Experiments}
\label{Experiments}
In this section, detailed network settings of our end-to-end C-HRNNs are firstly introduced. Next, C-HRNNs are compared with other popular methods on four challenging object/scene image classification benchmarks: ILSVRC 2012 \cite{deng2009imagenet}, Places 205 \cite{zhou2014learning}, SUN 397 \cite{xiao2010sun}, and MIT indoor \cite{quattoni2009recognizing}. Afterwards, effectiveness of different modules of C-HRNNs is analyzed, C-HSRN and C-HLSTM are compared in detail.

\subsection{Experimental Settings}
\label{Experimental settings}

\begin{table*}[!tp]\addtolength{\tabcolsep}{-1.3pt}
	\begin{tabular}{|c|c|c|c|c|c|c|c|c|}
		\hline
		Models & conv1                                                                                         & conv2                                                                                     & conv3                                                                           & conv4                                                                          & conv5                                                                                                                 & hrnn6                                                                                                                & fc7                                                            & fc8                                                            \\ 
		\hline
		\begin{tabular}[c]{@{}c@{}} Alex-net\\\cite{krizhevsky2012imagenet} \end{tabular}   & \begin{tabular}[c]{@{}c@{}}96x11x11\\ st. 4, pad 0\\ LRN, x2 pool\\ map 27x27\end{tabular} & \begin{tabular}[c]{@{}c@{}}256x5x5\\ st. 1, pad 2\\ LRN, x2 pool\\ map 13x13\end{tabular} & \begin{tabular}[c]{@{}c@{}}384x3x3\\ st. 1, pad 1\\ -\\ map 13x13\end{tabular}  & \begin{tabular}[c]{@{}c@{}}384x3x3\\ st. 1, pad 1\\ -\\ map 13x13\end{tabular} & \begin{tabular}[c]{@{}c@{}}256x3x3 \\ st. 1, pad 1\\ x2 pool\\ map 6x6\end{tabular}                                   & -                                                                                                                    & \begin{tabular}[c]{@{}c@{}}4096\\ drop-\\ out 0.5\end{tabular} & \begin{tabular}[c]{@{}c@{}}4096\\ drop-\\ out 0.5\end{tabular} \\ 
		\hline
		\begin{tabular}[c]{@{}c@{}} SPP-net\\\cite{he2014spatial} \end{tabular}    & \begin{tabular}[c]{@{}c@{}}96x7x7\\ st. 2, pad 1\\ LRN, x2 pool\\ map 55x55\end{tabular}      & \begin{tabular}[c]{@{}c@{}}256x5x5\\ st. 2, pad 0\\ LRN, x2 pool\\ map 13x13\end{tabular} & \begin{tabular}[c]{@{}c@{}}384x3x3\\ st. 1, pad 1\\ -\\ map 13x13\end{tabular}  & \begin{tabular}[c]{@{}c@{}}384x3x3\\ st. 1, pad 1\\ -\\ map 13x13\end{tabular} & \begin{tabular}[c]{@{}c@{}}256x3x3\\ st. 1, pad 1\\ \{x2, x4, x7, x13\}pool\\ map \{6x6, 3x3, 2x2, 1x1\}\end{tabular} & -                                                                                                                    & \begin{tabular}[c]{@{}c@{}}4096\\ drop-\\ out 0.5\end{tabular} & \begin{tabular}[c]{@{}c@{}}4096\\ drop-\\ out 0.5\end{tabular} \\ \hline
		C-HRNNs    & \begin{tabular}[c]{@{}c@{}}96x7x7\\ st. 2, pad 1\\ LRN, x2 pool\\ map 55x55\end{tabular}      & \begin{tabular}[c]{@{}c@{}}256x5x5\\ st. 2, pad 0\\ LRN, x2 pool\\ map 13x13\end{tabular} & \begin{tabular}[c]{@{}c@{}}384x3x3\\ st. 1, pad 1 \\ -\\ map 13x13\end{tabular} & \begin{tabular}[c]{@{}c@{}}384x3x3\\ st. 1, pad 1\\ -\\ map 13x13\end{tabular} & \begin{tabular}[c]{@{}c@{}}256x3x3\\ st. 1, pad 1\\ \{x2, x4, x7, x13\}pool\\ map \{6x6, 3x3, 2x2, 1x1\}\end{tabular} & \begin{tabular}[c]{@{}c@{}}\{36, 9, 4, 1\}x256x1x1\\ st.1, dropout 0.5\\ -\\ map \{6x6, 3x3, 2x2, 1x1\}\end{tabular} & \begin{tabular}[c]{@{}c@{}}4096\\ drop-\\ out 0.5\end{tabular} & \begin{tabular}[c]{@{}c@{}}4096\\ drop-\\ out 0.5\end{tabular} \\ \hline
	\end{tabular}
	\caption{Network structures. C-HSRN and C-HLSTM share the same structure, which are indicated as C-HRNNs.}
	\label{Table-C-HRNN-structure}
\end{table*}


Following the default data prepossessing settings in Caffe \cite{jia2014caffe}, all images are resized to $256\times{}256$ pixels and subtracted by the pixel mean. For training images, 10 sub-crops of size $224\times{}224$ (1 center, 4 corners, and horizontal flips) are extracted. In the remaining part of this section, if not specified, the results are the Top 1 accuracy (or error rates) tested with center crop by using a single model.

As shown in Table \ref{Table-C-HRNN-structure}, detailed layer structures of the baseline deep nets (Alex-net \cite{krizhevsky2012imagenet}, SPP-net \cite{he2014spatial}) and C-HRNNs are given. 

Comparing with SPP-net, our C-HRNNs has the same first five convolutional layers: $96(7\times{}7)$, $256(5\times{}5)$, $384(3\times{}3)$, $384(3\times{}3)$, and $256(3\times{}3)$ respectively.  Strides of the first two layers are 2, and the rest are 1. Following each of the first, second and fifth convolutional layers, there is a max pooling layer with kernel size of $3\times{}3$, and stride of 2. Finally, size of output feature maps of the fifth CNN layer is $256\times{}6\times{}6$ (number of channels $\times$ height $\times$ width). Similar to SPP-net, we pool the feature maps into four scales, and achieve response maps with size of $\{6\times{}6, 3\times{}3, 2\times{}2, 1\times{}1\}$. 

Different from all of these baseline networks, our C-HRNNs introduce hierarchical recurrent layers (hrnn6 as shown in Table \ref{Table-C-HRNN-structure}). For the hierarchical recurrent layers, we process three scale spatial RNN layers with size of $\{6\times{}6, 3\times{}3, 2\times{}2\}$ and one global $1\times{}1$ pooling layer, and build cross scale connections among all these four scales. The corresponding numbers of image regions of the four scales are 36, 9, 4, and 1 respectively, and each region is represented as a 256-dimensional feature vector (number of channels in the fifth layer CNN). For each RNN layer, sizes of the transformation matrices (HSRN: $W_{hh}^{r}$, $W_{hh}^{c}$, $W_{xh}^{}$, and $W_{jl}^{}$, refer to Equation \ref{HRNNquad_form_se}; HLSTM: $W_{hp}^{r}$, $W_{hp}^{c}$, $W_{xp}^{}$, and $W_{jl}^{}$, refer to Equation \ref{HLSTMquad_form_se}) in the four directional ``2D sequences" are $256\times{}256$.

To show the performance gain of introducing spatial and scale dependencies separately, we introduce an intermediate network called convolutional multi-scale recurrent neural networks (C-MRNNs), which only considers spatial dependencies in multiple scales, and ignores the scale dependencies. Specifically, convolutional multi-scale simple recurrent neural network (C-MSRN) and convolutional multi-scale long-short term memory neural network (C-MLSTM) are tested in the experiments. Furthermore, when all the hidden-hidden weights $W_{hh}$ in C-MSRN are set to 0, and the input-hidden weights $W_{ih}$ are identity matrices, C-MSRN degenerates to SPP-net. 

For the fully connected layers, the number of output units of both two layers is 4096, and each of them is applied dropout at the rate of 0.5.

The training batch size is 256, learning rate starts from 0.01 and it is divided by 10 when the accuracy stops increasing, and the weight of momentum is 0.9. All the experiments are run on Caffe \cite{jia2014caffe} with a single NVIDIA Tesla K40 GPU.

\subsection{Experimental Results}

\subsubsection{Experimental Results on ILSVRC 2012}
\begin{table*}[!tp] \addtolength{\tabcolsep}{9.5pt}
	\begin{center}
		\begin{tabular}{|l|c|c|c|c|}
			\hline
			Methods  & test scales & test views & Top 1 val & Top 5 val \\ 
			\hline\hline
			MOP-CNN \cite{gong2014multi} (max pooling)   & 3   & 101  & 44.12\%  & -   \\ 
			MOP-CNN \cite{gong2014multi} (VLAD pooling)  & 3   & 101  & 42.07\%  & -   \\ 
			\hline
			SPP-net \cite{he2014spatial}   & 1   & 1  & 38.21\%  &  -   \\ 
			C-MSRN  & 1   & 1  &  36.90\%  &  -   \\ 
			C-HSRN  & 1   & 1  & \textbf{36.38\%}  &  -   \\ 
			C-MLSTM  & 1   & 1  &  36.01\%  &  -   \\ 
			C-HLSTM  & 1   & 1  & \textbf{35.85\%}  &  -   \\ 
			\hline\hline
			Alex-net \cite{krizhevsky2012imagenet} & 1  & 10   & 40.7\%  & 18.2\%  \\ 
			ZF-net \cite{zeiler2013visualizing}  & 1   & 10   & 38.4\%  & 16.5\%   \\
			Overfeat\cite{sermanet2013overfeat}   & 1   & 10  & 35.6\%   & 14.7\%   \\ 
			\hline
			SPP-net \cite{he2014spatial} & 1   & 10  &  36.2\% &  14.9\%\\ 
			C-MSRN  & 1   & 10  & 35.2\%  &  14.0\%  \\ 
			C-HSRN  & 1   & 10  & \textbf{34.8\%}  &  \textbf{13.7\%}   \\ 
			C-MLSTM  & 1   & 10  & 34.5\%  &  13.5\%  \\ 
			C-HLSTM  & 1   & 10  & \textbf{34.3\%}  &  \textbf{13.4\%}   \\
			\hline
		\end{tabular}
	\end{center}
	\caption{Comparison of error rates on the ILSVRC 2012 validation set.}
	\label{Table-ILSVRC2012}
\end{table*}

ImageNet Large-Scale Visual Recognition Challenge (ILSVRC) dataset \cite{deng2009imagenet} is one of the most challenging and popular large-scale object image classification datasets. ILSVRC 2012 contains 1.2 million training images and 50,000 validation images (50 per class), and they belong to 1000 object categories.

In the upper part of Table \ref{Table-ILSVRC2012}, we compare C-HRNNs with SPP-net\cite{he2014spatial}, which encodes the spatial information by using spatial pyramid pooling. Based on their released model\footnote{\url{https://github.com/ShaoqingRen/SPP_net}}, we can only achieve 41.47\% top-1 error rate with one testing view. Therefore, we further tune the model with the settings in Section \ref{Experimental settings}, and finally achieve 38.21\% (reported 38.01\%) for SPP-net. The performance gap might be caused by the different training settings (when preprocess images, SPP-net keeps the original image aspect ratio, while the standard Caffe \cite{jia2014caffe} does not). C-HRNNs and SPP-net use the same convolutional and fully connected layer settings, except SPP-net directly applied $\{6\times{}6, 3\times{}3, 2\times{}2, 1\times{}1\}$ spatial pyramid pooling after the fifth convolutional layer, while our C-HRNNs model spatial dependencies with RNN for each scale, and models scale dependencies across different scales. 

For SRN models, comparing with SPP-net, C-MSRN brings 1.31\% Top 1 error rate decrease, which indicates the benefit of modeling spatial dependencies. After integrating the scale dependencies, C-HSRN is 1.83\% better than SPP-net. Thus, both encoding spatial and scale dependencies can help to generate better image representations. 

In LSTM models, performance improvement introduced by modeling spatial and scale dependencies can also be observed. Different from SRN models, LSTM models are able to keep longer-term memory of image region ``2D sequences". Comparing with SPP-net, C-HLSTM is 2.36\% better. C-MLSTM gets 36.01\%, which is better than C-MSRN, and C-HLSTM (35.85\%) also works better than C-HSRN. But the performance gap between C-HLSTM and C-HSRN (0.53\%) is less than the one between C-MLSTM and C-MSRN (0.89\%). The reason should be that the introduced scale dependencies from higher scales indirectly extend the long-term ability of C-MSRN, and indent the gap between SRN and LSTM models. 

We also compare with another spatial statistics based CNN method MOP-CNN \cite{gong2014multi}, which directly uses the Caffe CNN \cite{jia2014caffe} to densely extract features from three-scale image patches, and use VLAD pooling to generate global representations. The performance gap indicates that our way of encoding spatial and scale information is more effective.

In the lower part of Table \ref{Table-ILSVRC2012}, C-HRNNs are compared with other deep neural networks with the most general settings: 10 testing views, comparing Top 1 and Top 5 error rates. Outstanding performances of our C-HRNNs indicate that besides going deeper and wider, RNN is another promising way to increase the image representation power of neural networks.

\subsubsection{Experimental Results on Places 205}

\begin{table}[!tp] \addtolength{\tabcolsep}{8pt}
	\begin{center}
		\begin{tabular}{|l|c|c|c|}
			\hline
			Methods & Top 1 val & Top 5 val \\ 
			\hline\hline
			Alex-net \cite{krizhevsky2012imagenet, zhoulearning} & 50.06\% & 80.51\%\\
			SPP-net \cite{he2014spatial}  & 51.57\% & 81.88\%\\
			C-MSRN  &  52.70\% & 82.75\% \\
			C-HSRN &  \textbf{53.16\%} & \textbf{83.07\%} \\
			C-MLSTM  &  53.75\% & 83.36\% \\
			C-HLSTM &  \textbf{53.91\%} & \textbf{83.48\%} \\
			\hline
		\end{tabular}
	\end{center}
	\caption{Comparison of accuracy on Places 205.}
	\label{Table-Places205}
\end{table}

Places 205 dataset \cite{zhoulearning} is currently the largest scene categorization dataset, which has just been released at the end of 2014. Different from ILSVRC 2012, it focuses on scene images rather than object centric ones. It has 2.5 million training images from 205 scene categories, which is twice the size of ILSVRC 2012, and much more challenging. There are 20,500 images (100 per category) in the validation set. 


As shown in Table \ref{Table-Places205}, C-HLSTM update the state-of-the-art on Places 205 (previous best result was 50.06\% achieved by Alex-net) with the accuracy of 53.91\%. When only introducing spatial dependencies, C-MSRN and C-MLSTM outperform SPP-net by 1.13\% and 2.18\% respectively. Further integrating scale dependences, C-HSRN and C-HLSTM bring 1.59\% and 2.34\% improvements respectively.

\begin{table}[!tp] \addtolength{\tabcolsep}{10.5pt}
	\begin{center}
		\begin{tabular}{|l|c|}
			\hline
			Methods & Accuracy\\
			\hline\hline
			MOP-CNN \cite{gong2014multi} (max pooling) & 48.50\% \\
			MOP-CNN \cite{gong2014multi} (VLAD pooling) & \textbf{51.98\%} \\
			Alex-net \cite{krizhevsky2012imagenet} (ILSVRC ft) & 44.42\%\\
			Alex-net \cite{krizhevsky2012imagenet} (Places ft) & 54.55\%\\
			\hline
			SPP-net \cite{he2014spatial} (ILSVRC ft) & 49.02\%\\
			C-MSRN (ILSVRC ft) & 51.76\%\\
			C-HSRN (ILSVRC ft) & \textbf{52.59\%}\\
			C-MLSTM (ILSVRC ft) & 52.67\%\\
			C-HLSTM (ILSVRC ft) & \textbf{52.78\%}\\
			\hline 
			SPP-net \cite{he2014spatial} (Places ft) & 57.23\%\\
			C-MSRN (Places ft) & 59.32\%\\
			C-HSRN (Places ft) & \textbf{59.90\%}\\
			C-MLSTM (Places ft) & 60.08\%\\
			C-HLSTM (Places ft) & \textbf{60.34\%}\\
			\hline\hline
			Xiao et al.\cite{xiao2010sun} & 38.00\%\\
			IFV \cite{sanchez2013image} & 47.20\%\\
			MTL-SDCA \cite{lapinscalable} & 49.50\%\\
			\hline
		\end{tabular}
	\end{center}
	\caption{Comparison of accuracy on SUN 397.{\protect\footnotemark[2]}}
	\label{Table-SUN397}
\end{table}

\subsubsection{Experimental Results on SUN 397}
SUN 397 \cite{xiao2010sun} is another popular large-scale scene image recognition benchmark. There are 100,000 images from 397 scene classes in total. The general splittings in \cite{xiao2010sun} are used here, in which, there are 50 images per class for training, and 50 images per class for testing. Since the number of training images is too small (20,000), we introduce the models pre-trained on ILSVRC 2012 and Places 205, and use the training images from SUN 397 to fine-tune the network. We also increase the learning rates of the HRNN layers (10 times higher than the other layers), and aim to focus more on spatial dependencies specifically exist in SUN 397. 

As shown in the upper part of Table \ref{Table-SUN397}, our C-HRNNs performs better than existing CNNs. After fine-tuning on SUN 397, C-HRNNs are able to learn more data adaptive spatial dependencies and significantly outperform SPP-net: 1) Based on models pre-trained on ILSVRC, the performance gains of C-HSRN and C-HLSTM are 3.57\% and 3.76\% respectively; 2) Based on models pre-trained on Places, the accuracy improvements of C-HSRN and C-HLSTM are 2.67\% and 3.11\% correspondingly. 

When applying fine-tuning based on Places 205, C-HLSTM achieves the state-of-the-art on the SUN 397 with the accuracy of 60.34\%, which outperforms the previous best result (MOP-CNN 51.98\%) by 8.36\%. Another observation is that the performances of the fine-tuned models based on Places 205 consistently perform better than the ones based on ILSVRC 2012. The reason should be that both Places 205 and SUN 397 are scene datasets, their domain gap is smaller than the gap between ILSVRC 2012 (object dataset) and SUN 397. 

The lower part of Table \ref{Table-SUN397} shows the traditional state-of-the-art shallow methods. Most of these works heavily depend on combining multiple densely extracted hand-crafted features, and the image level representations are usually very high-dimensional. Another drawback of these methods is that the testing procedures are generally very time consuming, since the feature extraction steps are slow. Comparing with them, our C-HRNNs perform much better with much less computational cost in testing, and much lower-dimensional features.

\begin{table}[!tp] \addtolength{\tabcolsep}{10.5pt} 
	\begin{center}
		\begin{tabular}{|l|c|}
			\hline
			Methods & Accuracy\\
			\hline\hline
			MOP-CNN \cite{gong2014multi} (max pooling) & 64.85\% \\
			MOP-CNN \cite{gong2014multi} (VLAD pooling) & \textbf{68.88\%} \\
			Alex-net \cite{krizhevsky2012imagenet} (ILSVRC ft) & 61.57\%\\
			Alex-net \cite{krizhevsky2012imagenet} (Places ft) & 68.24\%\\
			\hline
			SPP-net \cite{he2014spatial} (ILSVRC ft) & 66.32\%\\
			C-MSRN (ILSVRC ft) & 68.28\%\\
			C-HSRN (ILSVRC ft) & \textbf{68.88\%}\\
			C-MLSTM (ILSVRC ft) & 69.18\%\\
			C-HLSTM (ILSVRC ft) & \textbf{69.25\%}\\
			\hline 

			SPP-net \cite{he2014spatial} (Places ft) & 72.09\%\\
			C-MSRN (Places ft) & 74.18\%\\
			C-HSRN (Places ft) & \textbf{74.85\%}\\
			C-MLSTM (Places ft) & 75.30\%\\
			C-HLSTM (Places ft) & \textbf{75.67\%}\\
			\hline\hline
			Object Bank \cite{li2010object} & 37.60\%\\
			Visual Concepts \cite{liharvesting} &  46.40\% \\
			MMDL \cite{wangmax} & 50.15\% \\
			IFV \cite{juneja2013blocks} & 60.77\% \\
			MLrep + IFV \cite{doersch2013mid} & 66.87\%\\
			ISPR + IFV \cite{linlearning} & 68.50\%\\
			\hline
		\end{tabular}
	\end{center}
	\caption{Comparison of accuracy on MIT indoor.{\protect\footnotemark[2]}}
	\label{Table-67indoor}
\end{table}
\footnotetext[2]{ILSVRC ft and Places ft represent the models fine-tuned based on the models pre-trained on ILSVRC 2012 and Places 205 respectively. }

\subsubsection{Experimental Results on MIT Indoor}

MIT indoor \cite{quattoni2009recognizing} is very challenging scene image classification benchmarks. This dataset focuses on indoor scene scenarios, which usually contains lots of objects, and has larger variations. There are 67 different scene scenarios in MIT indoor in total, and the widely used splitting provided by \cite{quattoni2009recognizing} are applied in our experiments. In each class, around 80 training images, and around 20 testing images are selected. Because of the limitation of dataset size, we also utilize the pre-train models on ILSVRC 2012 and Places 205, and do fine-tuning. For the other baseline deep neural networks, such fine-tuning is also applied.

As shown in the upper part of Table \ref{Table-67indoor}, C-HRNNs are able to outperform the other deep neural nets with obvious gaps. Comparing with the state-of-the art MOP-CNN, our C-HLSTM achieves the accuracy of 75.67\%, which is 6.79\% better. Comparing with SPP-net: 1) Based on models pre-trained on ILSVRC, the performance gains of C-HSRN and C-HLSTM are 2.56\% and 2.93\% respectively; 2) Based on models pre-trained on Places, the accuracy improvements of C-HSRN and C-HLSTM are 2.76\% and 3.58\% respectively. 

In the lower part of Table \ref{Table-67indoor}, results of the state-of-art traditional shallow methods are given. Although very powerful on MIT Indoor, these methods cost much more computation power to perform middle-level patch searching and clustering, the feature dimensions are relatively high, and most of them can hardly be applied on large-scale benchmarks. In contrast, our C-HRNNs are end-to-end feature learning frameworks with 4096-dimensional output features, and C-HRNNs can easily handle large-scale data.

\subsection{Analysis of C-HRNNs}
In this subsection, we will analyze the effectiveness of our C-HRNNs from different perspectives. 

\subsubsection{C-HRNNs Visualization}
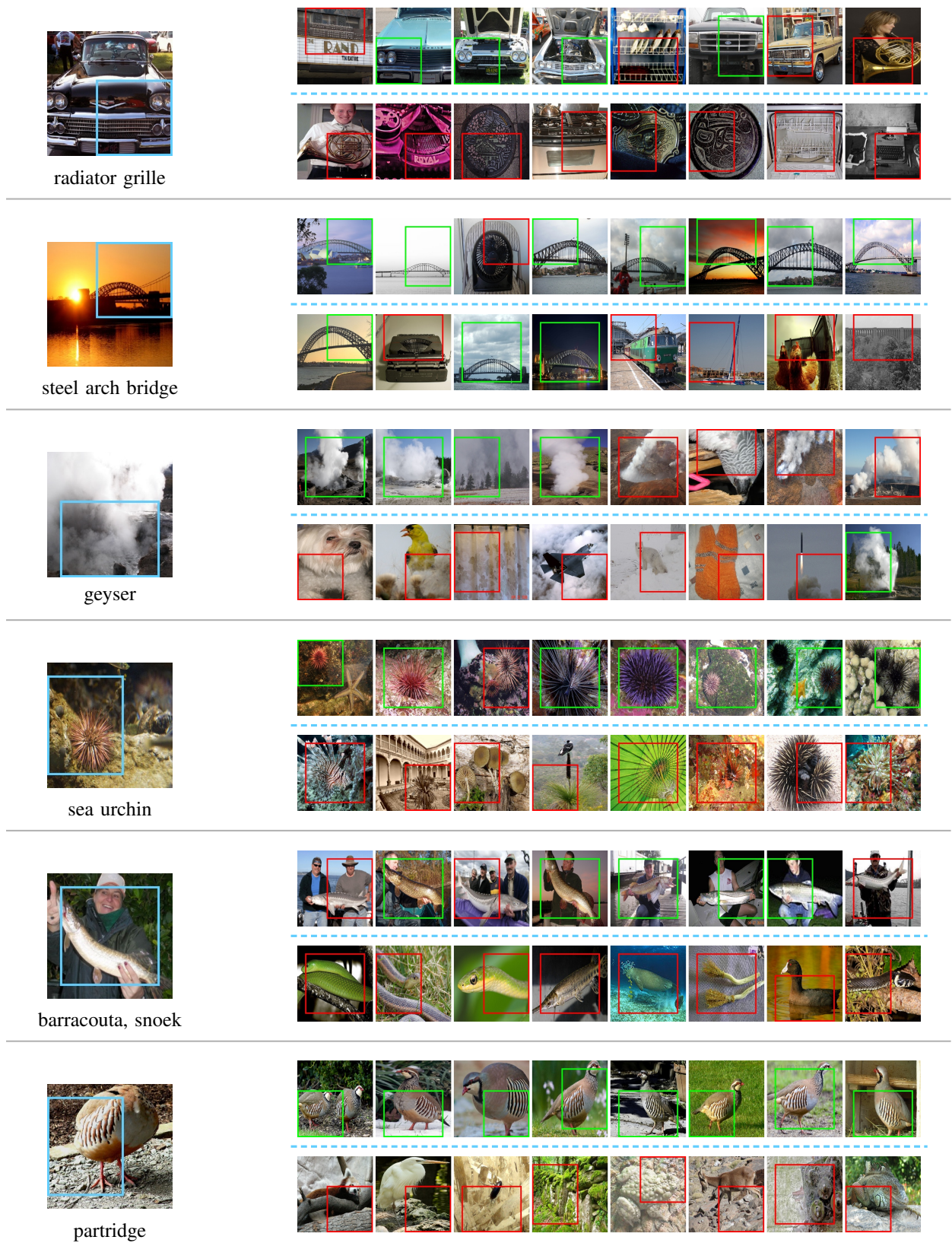
\begin{figure*}
	\newcommand{\compareimage}[4]{%
		\node (CHRNN#1_region)[on chain=1] {
			\includegraphics[width=0.6\linewidth]{figures/Visualization/remove_subimage/#1_#2_#3_C-HRNN.jpg}
		};
		\node (sppnet#1_region)[on chain=1] {
			\includegraphics[width=0.6\linewidth]{figures/Visualization/remove_subimage/#1_#2_#3_sppnet.jpg}
		};
		\draw [splitter, color=region]
			($(CHRNN#1_region.west)!.5!(sppnet#1_region.west)$) --
			($(CHRNN#1_region.east)!.5!(sppnet#1_region.east)$);
		\node (original#1) [original] at 
			($(CHRNN#1_region.west)!.5!(sppnet#1_region.west)$)
			{\includegraphics[width=0.12\linewidth]{figures/Visualization/remove_subimage/#1_#2_#3_original.jpg}};
		\node [below,text width=11em, text centered] at (original#1.south) {#4};
		\node [minimum width=2em, anchor=west] at
			($(CHRNN#1_region.east) + (1em, 0)$) {};
	}
	\newcommand{\drawsplitter}{
		\node (dummy)[on chain = 1, space]{};
		\draw [category_splitter, color = black!30]
		($(dummy) - (30em, 0)$) --
		($(dummy) + (17em, 0)$);
	}
	\begin{center}
	\begin{tikzpicture}
		\definecolor{region}{HTML}{60CEFF} 
		\tikzstyle{splitter} = [draw, very thick, densely dashed]
		\tikzstyle{original} = [xshift=-9em]
		\tikzstyle{space}=[minimum height = 1ex]
		\tikzstyle{category_splitter} = [draw, thick]
		\begin{scope}[start chain=1 going below,node distance=1mm]
			\compareimage{00009299}{581}{3_08}{radiator grille};
			\drawsplitter;
			\compareimage{00011262}{821}{3_02}{steel arch bridge};
			\drawsplitter;
			\compareimage{00034450}{974}{3_07}{geyser};
			\drawsplitter;
			\compareimage{00008817}{328}{3_03}{sea urchin};
			\drawsplitter;
			\compareimage{00037328}{389}{3_04}{barracouta, snoek};
			\drawsplitter;
			\compareimage{00040858}{086}{3_03}{partridge};
		\end{scope}
	\end{tikzpicture}
	\end{center}
	
	\caption{Visualization of the hrnn6 (C-HLSTM)/conv5 (SPP-net) layer features. The colored rectangles mark the receptive fields of each image region. First column on the left: testing image regions (blue rectangles) at scale 3x3. For each testing region, the regions on the right are the top 8 nearest neighbors searched by using C-HLSTM hrnn6 features (top row), and SPP-net conv5 features (bottom row). Green rectangles are the nearest neighbors with the same labels as the testing image region, red rectangles are the ones with the wrong labels. (Best viewed in color)}
	\label{Visualization}
\end{figure*}


Firstly, patterns learned by the hrnn6 layer (refer to Table \ref{Table-C-HRNN-structure}) of C-HLSTM are visualized in  Figure \ref{Visualization}. On the left part of Figure \ref{Visualization}, six testing image region on the \{3x3\} scale (refer to Section \ref{Experimental settings}) are given, and the receptive field of each region is highlighted with blue box in the original image. On the right part of Figure \ref{Visualization}, the top 8 nearest neighbors of each testing image region are shown. These nearest neighbors are searched from all the local region features extracted from training images, and measured by $\chi^2$ distance. For every two rows, the top row is the nearest neighbors searched by utilizing C-HLSTM hrnn6 layer features, and the bottom row is the results of using SPP-net conv5 layer features.

Comparing the visualization results of our C-HLSTM with the SPP-net, we can observe obvious better local image region representations. Take the first testing image region as an example, it is the right bottom area in the ``radiator grille" image. By using our C-HLSTM, this region is more likely to be represented as the ``radiator grille" from the same or different car models, and less likely to be mismatched to similar patterns from other unrelated classes. Take the last testing image ``partridge" as another example. The testing region contains ``body of partridge" and the background ``gravel". For our C-HLSTM, contextual information has been taken into consideration, thus, this region is represented as the ``body of partridge". In contrast, the SPP-net wrongly focused on ``gravel", and missed the target object.  Similarly, better local region visualization results of C-HLSTM can be observed in other classes, such as man-made buildings like ``steel arch bridge", creatures like ``sea urchin" etc.

\subsubsection{C-HRNNs vs Modified CNNs}
Since C-HRNNs have more parameters than the original CNNs, we aim to quantitatively show whether the performance gain is from encoding contextual dependencies, or simply from increasing the number of parameters. 

For each HRNN scale, there are four directional ``2D sequences". In HSRN, each direction has three transformation matrices $W_{hh}^{r}$, $W_{hh}^{c}$, and $W_{xh}$; While in HLSTM, each direction has four gate functions, and in each gate, there are $W_{hp}^{r}$, $W_{hp}^{c}$, and $W_{xp}$. Thus, for each scale, there are 12 transformation matrices in HSRN, and 48 matrices in HLSTM. Furthermore, in both HSRN and HLSTM, there are 6 cross scale transformation matrices $W_{jl}$. Each of these matrices has the size of $256\times{}256$, which has the same number of parameters as one convolution layer with 256 kernels of size $1\times{}1\times{}256$. Thus, in the modified CNNs, each transformation matrix in HRNN is replaced with a convolution layer.

Testing on ILSVRC 2012, HSRN gets 36.38\% in error rate, while the modified CNN gets 37.73\%. Similarly, HLSTM gets 35.85\% in error rate, while the modified CNN gets 37.49\%. These obvious gaps indicate that RNNs are able to learn contextual dependencies which cannot be captured by CNNs.

\subsubsection{Effect of Number of Spatial Context Directions}
In C-HRNNs, four directional ``2D sequences" are employed, can they really learn complementary information to each other? 

In Table \ref{Table-RNN-direction}, performance of C-HRNNs with different directions of spatial context are given. The first four rows show the performances of using single directional ``2D sequence", and different direction performs similarly to each other. On the last three rows of Table \ref{Table-RNN-direction}, results of combination of two directions, and the complete four directions are given. Comparing the results of using two directions and single direction, improvements can be observed. When combining all four directions, the best performance can be achieved.


\begin{table}[h] 
\setlength{\tabcolsep}{10pt}
\begin{center}
\begin{tabular}{|l|l|l@{\hskip -7pt}|l|l|}
\cline{1-2}\cline{4-5}
Methods  & Error && Methods  & Error\\ 
\cline{1-2}\cline{4-5}
\vspace{-1em}\\
\cline{1-2}\cline{4-5}
C-HSRN$_1(\searrow)$ & 36.96\%  && C-HLSTM$_1(\searrow)$ & 36.39\%\\ 
C-HSRN$_1(\nwarrow)$ & 37.03\%  && C-HLSTM$_1(\nwarrow)$ & 36.46\%\\
C-HSRN$_1(\swarrow)$ & 36.95\%  && C-HLSTM$_1(\swarrow)$ & 36.45\%\\
C-HSRN$_1(\nearrow)$ & 36.89\%  && C-HLSTM$_1(\nearrow)$ & 36.50\%\\ 
C-HSRN$_2({\searrow\!\!\nwarrow})$ & 36.85\%  && C-HLSTM$_2({\searrow\!\!\nwarrow})$ & 35.97\%\\ 
C-HSRN$_2(\swarrow\!\!\nearrow)$ & 36.59\%  && C-HLSTM$_2(\swarrow\!\!\nearrow)$ & 36.03\%\\ 
C-HSRN$_4$ & 36.38\%  && C-HLSTM$_4$ & 35.85\%\\
\cline{1-2}\cline{4-5}
\cline{1-2} 
\end{tabular}
\end{center}
\caption{Error rates of applying C-HRNNs with different spatial contextual directions on ILSVRC 2012. C-HSRN$_1$ and C-HLSTM$_1$ use one directional HRNNs; C-HSRN$_2$ and C-HLSTM$_2$ utilize two directions; C-HSRN$_4$ and C-HLSTM$_4$ use all the four directions.}
\label{Table-RNN-direction}
\end{table}

\subsubsection{HRNNs Complexity}
Although very powerful, our HRNNs do not bring much extra computational burden or memory usage. 

There are three scale HRNN layers with spatial dependencies encoded: $6\times{}6$, $3\times{}3$, and $2\times{}2$, each of them has 12 transformation matrices in HSRN and 48 transformation matrices in HLSTM, and there are 6 hierarchical connections (3 from $1\times{}1$, 2 from $2\times{}2$, and 1 from $3\times{}3$). Thus, there are 42 transformation matrices in HSRN and 150 matrices in HLSTM in total, each one has size of $256\times{}256$. Thus, the HSRN layers have $2,752,512$ parameters, and the HLSTM layers have $9,830,400$ parameters. The number of parameters in HLSTM is almost four times HSRN, which make the HLSTM models be able to learn more complex patterns with the price of more computation resources. In contrast, in CNN, the fully connected layers have most of the network parameters, e.g. the second fully connected layer needs to learn a $4096\times{}4096$ weight matrix, which has $16,777,216$ parameters, comparing with which, our HRNN layers have much fewer parameters.

In terms of memory consumption, HRNN layers do not cost much extra memory except some intermediate hidden layer output, i.e. $h^{(r,c)}$ and gate units $p^{(r,c)}$ (only exist in HLSTM) for each image region, which are 256-dimensional vectors. While in CNN, the most memory consuming part is the first convolutional layer. In our setting, output of the first CNN layer has 1,161,600 dimensions, comparing with which, the HRNNs cost negligible memory to save intermediate data.

\subsubsection{C-HRNNs Success \& Failure Cases}
%

\begin{figure*}[!htb]
	\centering
	\newlength{\CompSubLen}
	\setlength{\CompSubLen}{0.13\textwidth}
	\newcommand{\Comparison}[5]{
		\begin{subfigure}[t]{\CompSubLen}
			\captionsetup{
				justification=raggedleft, singlelinecheck=false,
				position=top
			}
			\includegraphics[width=\textwidth]{#1}
			\caption*{\scriptsize\scalebox{0.85}{#2 (#3)}\\
				\scalebox{0.85}{#4 (#5)}}
		\end{subfigure}
	}
	\newcounter{partcounter}
	\newcommand{\DatasetCaption}[1]{%
		\stepcounter{partcounter}%
		\\\vspace{-1.2em}%
		\caption*{(\alph{partcounter}) Images from #1 dataset.\vspace{1em}}
	}
	\newcommand{\Splitter}{%
		\vspace{-3pt}\\%
		\makebox[\textwidth][s]{%
			- - - - - - - - - - - - - - - - - - - - - - - - - - - - - -
			- - - - - - - - - - - - - - - - - - - - - - - - - - - - - -
			- - - - - - - - - - - - - - - - - - - - - - - - - -}%
		\vspace{3pt}\\%
	}
	\Comparison{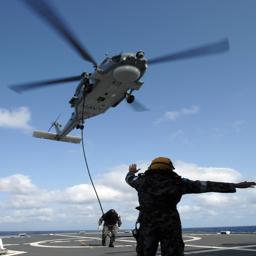}%
	{\textbf{landing deck}}{1.00}{windmill}{1.00}%
	\Comparison{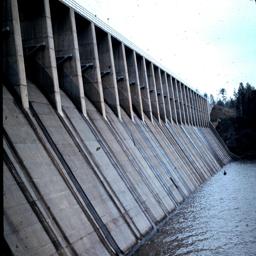}%
	{\textbf{dam}}{1.00}{aqueduct}{1.00}%
	\Comparison{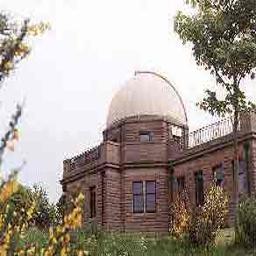}%
	{\textbf{observatory outdoor}}{1.00}{synagogue outdoor}{0.99}%
	\Comparison{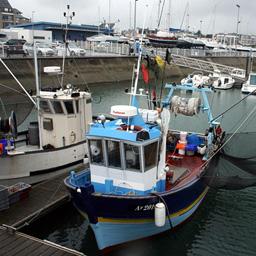}%
	{\textbf{harbor}}{1.00}{lido deck outdoor}{0.98}%
	\Comparison{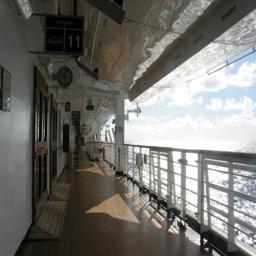}%
	{\textbf{promenade deck}}{1.00}{elevator shaft}{0.97}%
	\Comparison{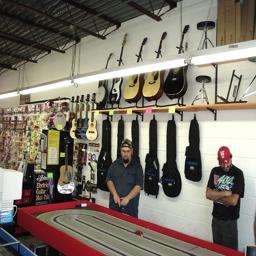}%
	{\textbf{music store}}{1.00}{gymn indoor}{0.95}%
	\Comparison{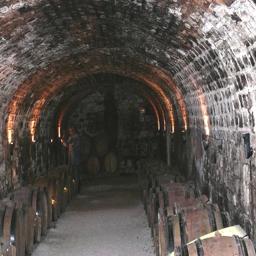}%
	{\hspace{-2ex}\textbf{wine cellar barrel}}{1.00}{catacomb}{0.94}%
	\Comparison{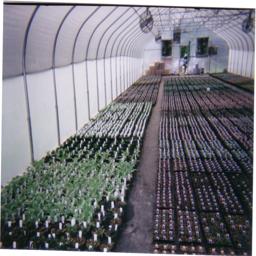}%
	{\textbf{greenhouse indoor}}{1.00}{fire escape}{0.93}%
	\Comparison{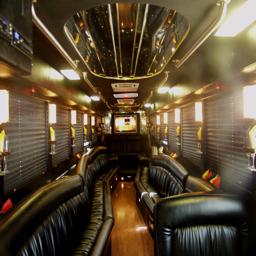}%
	{\textbf{limousine interior}}{1.00}{subway station platform}{0.93}%
	\Comparison{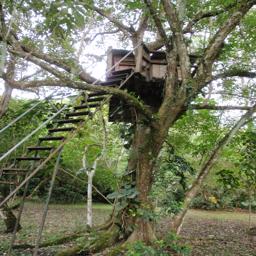}%
	{\textbf{tree house}}{1.00}{rainforest}{0.87}%
	\Comparison{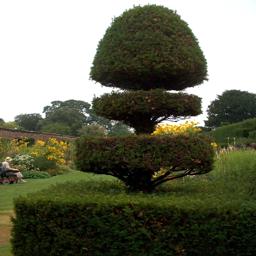}%
	{\textbf{topiary garden}}{1.00}{islet}{0.82}%
	\Comparison{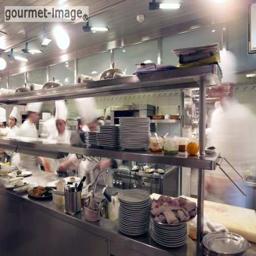}%
	{\textbf{restaurant kitchen}}{1.00}{shoe shop}{0.74}%
	\Comparison{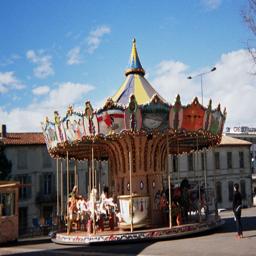}%
	{\textbf{carrousel}}{1.00}{temple east asia}{0.61}%
	\Comparison{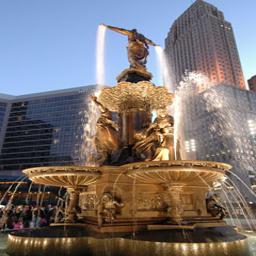}%
	{\textbf{fountain}}{1.00}{ruin}{0.50}%
	\Splitter%
	\Comparison{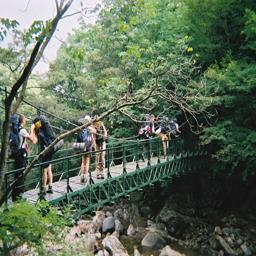}%
	{rainforest}{1.00}{\textbf{rope bridge}}{1.00}%
	\Comparison{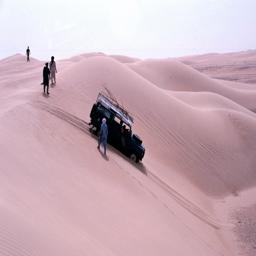}{raceway}{0.95}%
	{\textbf{desert sand}}{1.00}%
	\Comparison{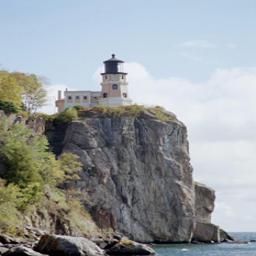}{cliff}{0.94}%
	{\textbf{lighthouse}}{1.00}%
	\Comparison{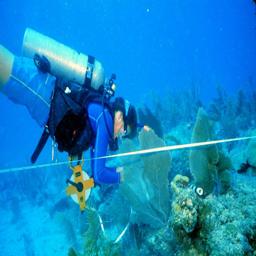}{aquarium}{0.64}%
	{\textbf{underwater coral reef}}{1.00}%
	\Comparison{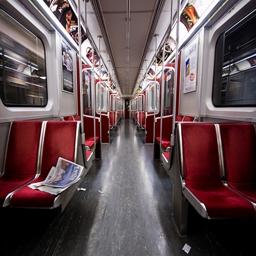}%
	{bus interior}{0.86}{\textbf{subway interior}}{1.00}%
	\Comparison{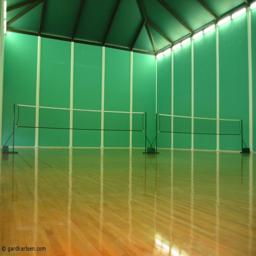}{shower}{0.60}%
	{\textbf{badminton court indoor}}{1.00}%
	\Comparison{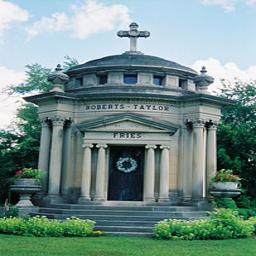}{mansion}{0.60}%
	{\textbf{mausoleum}}{1.00}%
	\caption{C-HLSTM vs SPP-net, result on SUN 397. The two rows above the dashed line: images misclassified by SPP-net, but correctly classified by C-HLSTM. The row below the dashed line: images correctly classified by SPP-net, but misclassified by C-HLSTM. Under each image, the first row shows the predicted label of using C-HLSTM, and the second row shows the predicted label of using SPP-net, the prediction confidence scores are shown in the bracket, and correct labels are in bold.}
	\label{C-HRNN_vs_SPP-net}
\end{figure*}

In Figure \ref{C-HRNN_vs_SPP-net}, the final classification results of using C-HLSTM and SPP-net \cite{he2014spatial} on SUN 397 (fined-tuned based on ILSVRC 2012 models) are visually compared. We show the images on which C-HLSTM leads to the highest accuracy improvement (the two rows above the dashed line), and the images on which C-HLSTM leads to the highest accuracy drop (the row below the dashed line). 

From the first two rows of Figure \ref{C-HRNN_vs_SPP-net}, we can clearly observe that the SPP-net focuses on predicting image regions, while ignoring the contextual information. For example, the label of the first image in the first row is ``landing deck", our C-HLSTM can correctly recognize it, while the SPP-net wrongly recognizes it as the ``windmill" with a very high confidence score. It's because SPP-net wrongly recognized the rotor blades of the helicopter as the windmill blades. In contrast, our C-HLSTM takes the context such as the body of helicopter and deck into consideration. Thus, our C-HLSTM works better when the local image regions are confusing, but contextual information can help to make better decisions.

In the third row of Figure \ref{C-HRNN_vs_SPP-net}, we observe some interesting results. For example, the first image of the third row is ``rope bridge", which is relatively small in the image, while the forest is more obvious. Thus, our C-HLSTM wrongly recognize it as ``rainforest". For the third image of the third row, the first word that comes to mind is cliff, while the ground truth label ``light house" just represents a small region on the ``cliff". Thus, our C-HLSTM makes mistakes when class labels are based on local regions rather than the global image.


\section{Conclusions}
In this manuscript, we propose an end-to-end deep learning framework to encode spatial and scale contextual dependencies in image representation, which is called C-HRNNs. In C-HRNNs, CNN layers are firstly utilize to extract middle-level representations for local image regions. Based on the CNN layer outputs, our proposed hierarchical recurrent layers are then applied to model the spatial dependencies among different image regions from the same scale, and the scale dependencies among image regions from different scales but at the same locations. In our proposed hierarchical recurrent neural networks, HSRN and HLSTM are introduced as two specific instances, which correspond to a fast recurrent model, and a sophisticated but more effective recurrent model respectively. By integrating CNN and HRNNs, our C-HRNNs show outstanding performances on image classification. 


\bibliographystyle{IEEEtran}
\bibliography{egbib}

\begin{thebibliography}{10}
\providecommand{\url}[1]{#1}
\csname url@samestyle\endcsname
\providecommand{\newblock}{\relax}
\providecommand{\bibinfo}[2]{#2}
\providecommand{\BIBentrySTDinterwordspacing}{\spaceskip=0pt\relax}
\providecommand{\BIBentryALTinterwordstretchfactor}{4}
\providecommand{\BIBentryALTinterwordspacing}{\spaceskip=\fontdimen2\font plus
\BIBentryALTinterwordstretchfactor\fontdimen3\font minus
  \fontdimen4\font\relax}
\providecommand{\BIBforeignlanguage}[2]{{%
\expandafter\ifx\csname l@#1\endcsname\relax
\typeout{** WARNING: IEEEtran.bst: No hyphenation pattern has been}%
\typeout{** loaded for the language `#1'. Using the pattern for}%
\typeout{** the default language instead.}%
\else
\language=\csname l@#1\endcsname
\fi
#2}}
\providecommand{\BIBdecl}{\relax}
\BIBdecl

\bibitem{le1990handwritten}
B.~B. Le~Cun, J.~Denker, D.~Henderson, R.~E. Howard, W.~Hubbard, and L.~D.
  Jackel, ``Handwritten digit recognition with a back-propagation network,'' in
  \emph{NIPS}, 1990.

\bibitem{krizhevsky2012imagenet}
A.~Krizhevsky, I.~Sutskever, and G.~Hinton, ``Imagenet classification with deep
  convolutional neural networks,'' in \emph{NIPS}, 2012.

\bibitem{zhou2014learning}
B.~Zhou, A.~Lapedriza, J.~Xiao, A.~Torralba, and A.~Oliva, ``Learning deep
  features for scene recognition using places database,'' in \emph{NIPS}, 2014.

\bibitem{gong2014multi}
Y.~Gong, L.~Wang, R.~Guo, and S.~Lazebnik, ``Multi-scale orderless pooling of
  deep convolutional activation features,'' in \emph{ECCV}, 2014.

\bibitem{chatfield2014return}
K.~Chatfield, K.~Simonyan, A.~Vedaldi, and A.~Zisserman, ``Return of the devil
  in the details: Delving deep into convolutional nets,'' in \emph{BMVC}, 2014.

\bibitem{szegedy2014going}
C.~Szegedy, W.~Liu, Y.~Jia, P.~Sermanet, S.~Reed, D.~Anguelov, D.~Erhan,
  V.~Vanhoucke, and A.~Rabinovich, ``Going deeper with convolutions,''
  \emph{arXiv preprint arXiv:1409.4842}, 2014.

\bibitem{simonyan2014very}
K.~Simonyan and A.~Zisserman, ``Very deep convolutional networks for
  large-scale image recognition,'' \emph{arXiv preprint arXiv:1409.1556}, 2014.

\bibitem{he2014spatial}
K.~He, X.~Zhang, S.~Ren, and J.~Sun, ``Spatial pyramid pooling in deep
  convolutional networks for visual recognition,'' in \emph{ECCV}, 2014.

\bibitem{girshick2014rcnn}
R.~Girshick, J.~Donahue, T.~Darrell, and J.~Malik, ``Rich feature hierarchies
  for accurate object detection and semantic segmentation,'' in \emph{CVPR},
  2014.

\bibitem{sermanet2013overfeat}
P.~Sermanet, D.~Eigen, X.~Zhang, M.~Mathieu, R.~Fergus, and Y.~LeCun,
  ``Overfeat: Integrated recognition, localization and detection using
  convolutional networks,'' \emph{arXiv preprint arXiv:1312.6229}, 2013.

\bibitem{ouyang2014deepid}
W.~Ouyang, P.~Luo, X.~Zeng, S.~Qiu, Y.~Tian, H.~Li, S.~Yang, Z.~Wang, Y.~Xiong,
  C.~Qian \emph{et~al.}, ``Deepid-net: multi-stage and deformable deep
  convolutional neural networks for object detection,'' \emph{arXiv preprint
  arXiv:1409.3505}, 2014.

\bibitem{taigman2014deepface}
Y.~Taigman, M.~Yang, M.~Ranzato, and L.~Wolf, ``Deepface: Closing the gap to
  human-level performance in face verification,'' in \emph{CVPR}, 2014.

\bibitem{sun2014deep}
Y.~Sun, X.~Wang, and X.~Tang, ``Deep learning face representation from
  predicting 10,000 classes,'' in \emph{CVPR}, 2014.

\bibitem{mikolov2012statistical}
T.~Mikolov, ``Statistical language models based on neural networks,'' Ph.D.
  dissertation, Brno University of Technology, 2012.

\bibitem{sutskever2013training}
I.~Sutskever, ``Training recurrent neural networks,'' Ph.D. dissertation,
  University of Toronto, 2013.

\bibitem{koutnik2014clockwork}
J.~Koutn{\'\i}k, K.~Greff, F.~Gomez, and J.~Schmidhuber, ``A clockwork rnn,''
  in \emph{ICML}, 2014.

\bibitem{graves2014towards}
A.~Graves and N.~Jaitly, ``Towards end-to-end speech recognition with recurrent
  neural networks,'' in \emph{ICML}, 2014.

\bibitem{sutskever2009recurrent}
I.~Sutskever, G.~E. Hinton, and G.~W. Taylor, ``The recurrent temporal
  restricted boltzmann machine,'' in \emph{NIPS}, 2009.

\bibitem{boulanger2012modeling}
N.~Boulanger-Lewandowski, Y.~Bengio, and P.~Vincent, ``Modeling temporal
  dependencies in high-dimensional sequences: Application to polyphonic music
  generation and transcription,'' in \emph{ICML}, 2012.

\bibitem{elman1990finding}
J.~L. Elman, ``Finding structure in time,'' \emph{Cognitive science}, vol.~14,
  no.~2, pp. 179--211, 1990.

\bibitem{jaeger2002tutorial}
H.~Jaeger, \emph{Tutorial on training recurrent neural networks, covering BPPT,
  RTRL, EKF and the ``echo state network" approach}.\hskip 1em plus 0.5em minus
  0.4em\relax GMD-Forschungszentrum Informationstechnik, 2002.

\bibitem{graves2009offline}
A.~Graves and J.~Schmidhuber, ``Offline handwriting recognition with
  multidimensional recurrent neural networks,'' in \emph{NIPS}, 2009.

\bibitem{oquab2014learning}
M.~Oquab, L.~Bottou, I.~Laptev, J.~Sivic \emph{et~al.}, ``Learning and
  transferring mid-level image representations using convolutional neural
  networks,'' in \emph{CVPR}, 2014.

\bibitem{donahue2013decaf}
J.~Donahue, Y.~Jia, O.~Vinyals, J.~Hoffman, N.~Zhang, E.~Tzeng, and T.~Darrell,
  ``Decaf: A deep convolutional activation feature for generic visual
  recognition,'' in \emph{ICML}, 2014.

\bibitem{hinton2006fast}
G.~E. Hinton, S.~Osindero, and Y.~W. Teh, ``A fast learning algorithm for deep
  belief nets,'' \emph{Neural computation}, 2006.

\bibitem{lee2009convolutional}
H.~Lee, R.~Grosse, R.~Ranganath, and A.~Y. Ng, ``Convolutional deep belief
  networks for scalable unsupervised learning of hierarchical
  representations,'' in \emph{ICML}, 2009.

\bibitem{nair20093d}
V.~Nair and G.~E. Hinton, ``3d object recognition with deep belief nets,'' in
  \emph{NIPS}, 2009.

\bibitem{hinton2006reducing}
G.~E. Hinton and R.~R. Salakhutdinov, ``Reducing the dimensionality of data
  with neural networks,'' \emph{Science}, vol. 313, no. 5786, pp. 504--507,
  2006.

\bibitem{yan2014modeling}
X.~Yan, H.~Chang, S.~Shan, and X.~Chen, ``Modeling video dynamics with deep
  dynencoder,'' in \emph{ECCV}, 2014.

\bibitem{zhang2014coarse}
J.~Zhang, S.~Shan, M.~Kan, and X.~Chen, ``Coarse-to-fine auto-encoder networks
  (cfan) for real-time face alignment,'' in \emph{ECCV}, 2014.

\bibitem{bengio2013generalized}
Y.~Bengio, L.~Yao, G.~Alain, and P.~Vincent, ``Generalized denoising
  auto-encoders as generative models,'' in \emph{NIPS}, 2013.

\bibitem{sutskever2011generating}
I.~Sutskever, J.~Martens, and G.~E. Hinton, ``Generating text with recurrent
  neural networks,'' in \emph{ICML}, 2011.

\bibitem{graves2013generating}
A.~Graves, ``Generating sequences with recurrent neural networks,'' \emph{arXiv
  preprint arXiv:1308.0850}, 2013.

\bibitem{pinheiro2014recurrent}
P.~Pinheiro and R.~Collobert, ``Recurrent convolutional neural networks for
  scene labeling,'' in \emph{ICML}, 2014.

\bibitem{rolfe2013discriminative}
J.~T. Rolfe and Y.~LeCun, ``Discriminative recurrent sparse auto-encoders,''
  \emph{arXiv preprint arXiv:1301.3775}, 2013.

\bibitem{eigen2013understanding}
D.~Eigen, J.~Rolfe, R.~Fergus, and Y.~LeCun, ``Understanding deep architectures
  using a recursive convolutional network,'' \emph{arXiv preprint
  arXiv:1312.1847}, 2013.

\bibitem{gregor2015draw}
K.~Gregor, I.~Danihelka, A.~Graves, and D.~Wierstra, ``Draw: A recurrent neural
  network for image generation,'' \emph{arXiv preprint arXiv:1502.04623}, 2015.

\bibitem{mnih2014recurrent}
V.~Mnih, N.~Heess, A.~Graves, and K.~Kavukcuoglu, ``Recurrent models of visual
  attention,'' in \emph{NIPS}, 2014, pp. 2204--2212.

\bibitem{donahue2014long}
J.~Donahue, L.~A. Hendricks, S.~Guadarrama, M.~Rohrbach, S.~Venugopalan,
  K.~Saenko, and T.~Darrell, ``Long-term recurrent convolutional networks for
  visual recognition and description,'' \emph{arXiv preprint arXiv:1411.4389},
  2014.

\bibitem{ng2015beyond}
J.~Y.-H. Ng, M.~Hausknecht, S.~Vijayanarasimhan, O.~Vinyals, R.~Monga, and
  G.~Toderici, ``Beyond short snippets: Deep networks for video
  classification,'' \emph{arXiv preprint arXiv:1503.08909}, 2015.

\bibitem{chen2014learning}
X.~Chen and C.~L. Zitnick, ``Learning a recurrent visual representation for
  image caption generation,'' \emph{arXiv preprint arXiv:1411.5654}, 2014.

\bibitem{karpathy2014deep}
A.~Karpathy and L.~Fei-Fei, ``Deep visual-semantic alignments for generating
  image descriptions,'' \emph{arXiv preprint arXiv:1412.2306}, 2014.

\bibitem{venugopalan2014translating}
S.~Venugopalan, H.~Xu, J.~Donahue, M.~Rohrbach, R.~Mooney, and K.~Saenko,
  ``Translating videos to natural language using deep recurrent neural
  networks,'' \emph{arXiv preprint arXiv:1412.4729}, 2014.

\bibitem{mao2014deep}
J.~Mao, W.~Xu, Y.~Yang, J.~Wang, and A.~Yuille, ``Deep captioning with
  multimodal recurrent neural networks (m-rnn),'' \emph{arXiv preprint
  arXiv:1412.6632}, 2014.

\bibitem{behnke2003hierarchical}
S.~Behnke, \emph{Hierarchical neural networks for image interpretation}.\hskip
  1em plus 0.5em minus 0.4em\relax Springer Science \& Business Media, 2003,
  vol. 2766.

\bibitem{visin2015renet}
F.~Visin, K.~Kastner, K.~Cho, M.~Matteucci, A.~Courville, and Y.~Bengio,
  ``Renet: A recurrent neural network based alternative to convolutional
  networks,'' \emph{arXiv preprint arXiv:1505.00393}, 2015.

\bibitem{zeiler2013visualizing}
M.~D. Zeiler and R.~Fergus, ``Visualizing and understanding convolutional
  neural networks,'' \emph{arXiv preprint arXiv:1311.2901}, 2013.

\bibitem{lowe2004distinctive}
D.~G. Lowe, ``Distinctive image features from scale-invariant keypoints,''
  \emph{International journal of computer vision}, vol.~60, no.~2, pp. 91--110,
  2004.

\bibitem{dalal2005histograms}
N.~Dalal and B.~Triggs, ``Histograms of oriented gradients for human
  detection,'' in \emph{CVPR}, 2005.

\bibitem{hochreiter1997long}
S.~Hochreiter and J.~Schmidhuber, ``Long short-term memory,'' \emph{Neural
  computation}, vol.~9, no.~8, pp. 1735--1780, 1997.

\bibitem{jia2014caffe}
Y.~Jia, E.~Shelhamer, J.~Donahue, S.~Karayev, J.~Long, R.~Girshick,
  S.~Guadarrama, and T.~Darrell, ``Caffe: Convolutional architecture for fast
  feature embedding,'' \emph{arXiv preprint arXiv:1408.5093}, 2014.

\bibitem{deng2009imagenet}
J.~Deng, W.~Dong, R.~Socher, L.~J. Li, K.~Li, and L.~Fei-Fei, ``Imagenet: A
  large-scale hierarchical image database,'' in \emph{CVPR}, 2009.

\bibitem{xiao2010sun}
J.~Xiao, J.~Hays, K.~A. Ehinger, A.~Oliva, and A.~Torralba, ``Sun database:
  Large-scale scene recognition from abbey to zoo,'' in \emph{CVPR}, 2010.

\bibitem{quattoni2009recognizing}
A.~Quattoni and A.~Torralba, ``Recognizing indoor scenes,'' in \emph{CVPR},
  2009.

\bibitem{zhoulearning}
B.~Zhou, A.~Lapedriza, J.~Xiao, A.~Torralba, and A.~Oliva, ``Learning deep
  features for scene recognition using places database,'' in \emph{NIPS}, 2014.

\bibitem{sanchez2013image}
J.~S{\'a}nchez, F.~Perronnin, T.~Mensink, and J.~Verbeek, ``Image
  classification with the fisher vector: Theory and practice,''
  \emph{International journal of computer vision}, vol. 105, no.~3, pp.
  222--245, 2013.

\bibitem{lapinscalable}
M.~Lapin, B.~Schiele, and M.~Hein, ``Scalable multitask representation learning
  for scene classification,'' in \emph{CVPR}, 2014.

\bibitem{li2010object}
L.-J. Li, H.~Su, L.~Fei-Fei, and E.~P. Xing, ``Object bank: A high-level image
  representation for scene classification \& semantic feature sparsification,''
  in \emph{NIPS}, 2010.

\bibitem{liharvesting}
Q.~Li, J.~Wu, and Z.~Tu, ``Harvesting mid-level visual concepts from
  large-scale internet images,'' in \emph{CVPR}, 2013.

\bibitem{wangmax}
X.~Wang, B.~Wang, X.~Bai, W.~Liu, and Z.~Tu, ``Max-margin multiple-instance
  dictionary learning,'' in \emph{ICML}, 2013.

\bibitem{juneja2013blocks}
M.~Juneja, A.~Vedaldi, C.~Jawahar, and A.~Zisserman, ``Blocks that shout:
  Distinctive parts for scene classification,'' in \emph{CVPR}, 2013.

\bibitem{doersch2013mid}
C.~Doersch, A.~Gupta, and A.~A. Efros, ``Mid-level visual element discovery as
  discriminative mode seeking,'' in \emph{NIPS}, 2013.

\bibitem{linlearning}
D.~Lin, C.~Lu, R.~Liao, and J.~Jia, ``Learning important spatial pooling
  regions for scene classification,'' in \emph{CVPR}, 2014.

\end{thebibliography}

%
%

%



\newcommand{\AuthorBio}[3]{%
\begin{IEEEbiography}[{%
	\includegraphics[width=1in, keepaspectratio]{figures/Bio/#2.jpg}%
}]{#1}
	#3
\end{IEEEbiography}%
}

\vfill


\end{document}